\documentclass[lettersize,journal]{IEEEtran}
\usepackage{amsmath,amssymb,amsfonts}
\usepackage{algorithmic}
\usepackage{algorithm}
\usepackage{array}
\usepackage[caption=false,font=normalsize,labelfont=sf,textfont=sf]{subfig}
\usepackage{textcomp}
\usepackage{stfloats}
\usepackage{url}
\usepackage{verbatim}
\usepackage{graphicx}
\usepackage{cite}
\usepackage{xspace}
\usepackage{url}
\usepackage[utf8]{inputenc}
\usepackage{tabularx}
\usepackage{booktabs}
\usepackage{xcolor}
\usepackage{tikz}
\usetikzlibrary{arrows.meta, positioning, calc, 3d, shapes.geometric}
\usepackage{tikz-3dplot}
\usepackage{caption} 
\usepackage{subcaption} 
\usepackage[colorlinks=true, linkcolor=red, citecolor=blue, urlcolor=blue]{hyperref}
\usepackage{makecell}   
\usepackage{ragged2e}
\usepackage{pgfplots}
\pgfplotsset{compat=1.18} 
\usetikzlibrary{arrows.meta, positioning, calc, 3d}

\definecolor{hallucRed}{RGB}{200, 0, 0}
\definecolor{correctGreen}{RGB}{0, 120, 0}

\newcommand{\Npairs}{84{,}631} 
\newcommand{\Ntest}{1{,}000}

\newcommand{\sdiv}{\textsc{SDIV}}
\definecolor{tpami_red}{RGB}{200, 0, 0}
\definecolor{tpami_green}{RGB}{0, 100, 0}
\newcommand{\halluc}[1]{\textcolor{tpami_red}{\textbf{#1}}}
\newcommand{\fact}[1]{\textcolor{tpami_green}{\textbf{#1}}}

\begin{document}

\title{Suppressing Prior-Comparison Hallucinations in Radiology Report Generation via Semantically Decoupled Latent Steering}

\author{Ao Li, Rui Liu, Mingjie Li, Sheng Liu, Lei Wang, \IEEEmembership{Senior~Member,~IEEE}, Xiaodan Liang, \IEEEmembership{Member,~IEEE}, Lina Yao, \IEEEmembership{Senior~Member,~IEEE}, Xiaojun Chang, \IEEEmembership{Senior~Member,~IEEE}, Lei Xing
\thanks{Ao Li and Lina Yao are with the University of New South Wales (email: \{ao.li, lina.yao\}@unsw.edu.au). Rui Liu is with the Australian Artificial Intelligence Institute, University of Technology Sydney (email: rui.liu-8@student.uts.edu.au). Mingjie Li, Sheng Liu and Lei Xing are with the Stanford University (email: \{lmj695, shengl, lei\}@stanford.edu). Lei Wang is with the School of Computing and Information Technology of University of Wollongong Australia (email: leiw@uow.edu.au). Xiaodan Liang is with the Sun Yat-sen University (email: liangxd9@mail.sysu.edu.cn). Xiaojun Chang is with the University of Science and Technology of China (email: xjchang@ustc.edu.cn).}
\thanks{Corresponding authors are Mingjie Li and Lei Xing. Code is available at \url{https://github.com/AoXan/SDLS-Radiology}.}
} 

\markboth{Under Review}%
{Li \MakeLowercase{\textit{et al.}}: Suppressing Prior-Comparison Hallucinations in Radiology Report Generation via Semantically Decoupled Latent Steering}


\maketitle

\begin{abstract}
Automated radiology report generation using vision-language models (VLMs) is limited by the risk of prior-comparison hallucination, where the model generates historical findings unsupported by the current study.
We address this challenge with a training-free, inference-time control framework termed Semantically Decoupled Latent Steering (SDLS).
Unlike generic activation steering, which often suffers from semantic entanglement, our approach constructs a semantic-free intervention vector via large language model (LLM)-driven semantic decomposition followed by $QR$-based orthogonalization.
This orthogonalization step is critical. It leverages geometric constraints to filter out the clinical semantics often entangled in standard principal component analysis (PCA) directions, ensuring that the steering vector targets only the ``historical comparison" axis.
We validate our method on the BiomedGPT foundation model, demonstrating that it overcomes the trade-off between hallucination suppression and clinical accuracy.
Extensive experiments on MIMIC-CXR, and zero-shot transfer evaluation on CheXpert Plus and IU-Xray, demonstrate the robustness of our approach.
Quantitative evaluations on MIMIC-CXR show that our approach significantly reduces the probability of historical hallucinations (FilBERT score decreases from 0.2373 to 0.1889) and improves clinical label fidelity (CheXpert macro-F1 increases from 0.2242 to 0.3208). Supplementary evaluations confirm that the structural integrity of the clinical narrative is maintained.
\end{abstract}

\begin{IEEEkeywords}
Medical report Generation, Hallucination Suppression, Latent Space Steering, Mechanistic Interpretability.
\end{IEEEkeywords}

\section{Introduction}
\label{sec:introduction}
\IEEEPARstart{R}{adiology} reports are expected to be faithful descriptions of current imaging findings.
However, automatic systems frequently hallucinate references to prior exams that do not exist, producing phrases like ``no interval change" or ``worsening opacity" even when processing a single study.
This phenomenon, termed \textit{prior-comparison hallucination}, undermines the temporal specificity of the report and acts as a barrier to clinical trust~\cite{ramesh2022filbert, BelislePipon2024}.
Such hallucinations often stem from the skewed distribution of large-scale medical corpora, where comparisons are ubiquitous, leading models to internalize strong statistical priors that override visual evidence~\cite{liu2024survey, zhang2024vision}.
This misalignment reflects a broader challenge in multimodal machine learning~\cite{baltrusaitis2019multimodal}, where the ``modality gap'' often compromises the interpretability and trustworthiness of deep neural networks~\cite{zhang2021visual}, a phenomenon identified as a critical safety risk in recent surveys of medical foundation models~\cite{kim2025medical}.

\begin{figure}[t]
    \centering
    \includegraphics[width=1\linewidth]{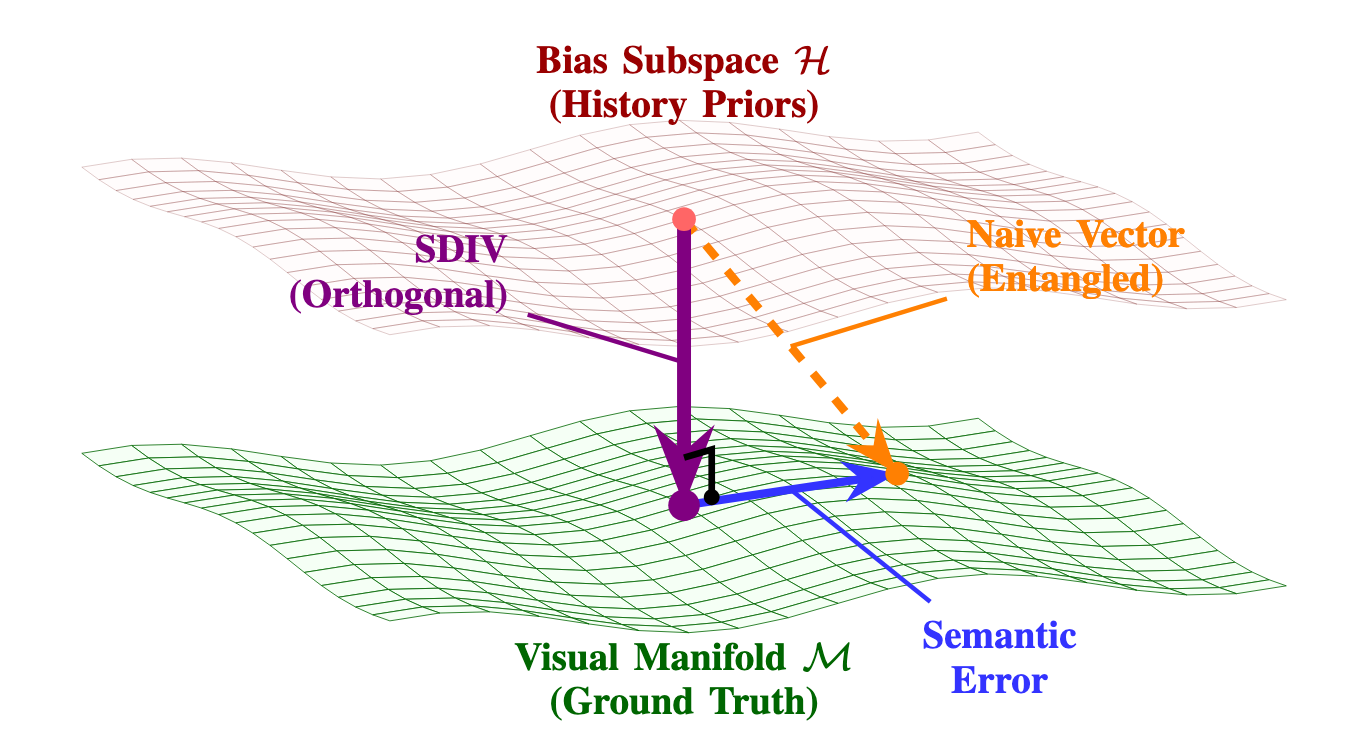}
    \caption{\textbf{Geometric Logic of SDLS.} 
The latent space consists of two conflicting components: the \textcolor{green!40!black}{\textbf{Visual Manifold}}, which encodes grounded image features representing the clinical ground truth (e.g., current evidence of `pleural effusion'), and the \textcolor{red!60!black}{\textbf{Bias Subspace}}, which captures statistical language priors (e.g., the tendency to generate `no interval change' regardless of visual evidence).
Standard steering produces an \textcolor{orange!100}{\textbf{Entangled Naive Vector}} (dashed) that inadvertently shifts representations on the manifold, creating a \textcolor{blue!80!white}{\textbf{Semantic Error}}.
Our \textcolor{violet}{\textbf{Semantically Decoupled Intervention Vector(SDIV)}} utilizes $QR$ decomposition to isolate the strictly \textbf{Orthogonal Component}, projecting representations vertically to remove bias without spatial distortion.}
    \label{fig:teaser_mesh}
\end{figure}

To mitigate this, the dominant strategy currently involves data-centric interventions: training classifiers to filter or re-write historical references, followed by retraining the generator on the cleaned corpus~\cite{ramesh2022filbert}.
While effective at reducing hallucination rates, this approach incurs significant operational costs, as retraining large foundation models is computationally expensive and must be repeated for every model update~\cite{pham2024towards}.
More critically, this method effectively impairs the model's longitudinal reasoning capabilities.
By removing all valid comparisons from the training distribution, the model permanently loses the ability to express valid temporal changes (e.g., ``stable'' or ``worsened'') even when legitimate historical context is provided.
It treats the symptom (dataset bias) but damages the model's general utility.

We advocate for a paradigm shift from expensive retraining to \textit{training-free mechanistic latent steering}. In this approach, we directly modulate the model's internal hidden states (activations) during inference, pushing the representation away from hallucination-prone directions.
Building on the principles of Representation Engineering (RepE)~\cite{zou2023representation}, this inference-time approach posits that high-level semantic concepts are encoded as linear directions in the model's activation space~\cite{saglam2025large, siddique2025shifting}.
Techniques such as \textit{Activation Addition}~\cite{turner2023actadd} and \textit{Function Vectors}~\cite{todd2023function} have successfully modulated attributes like sentiment or honesty in large language model (LLM) without parameter optimization.
Crucially, this allows for dynamic control: the steering can be applied or removed instantly depending on the availability of priors, avoiding the permanent capability loss of retraining.

However, applying these steering methods to radiology presents a unique challenge: the problem of \textit{semantic entanglement}.
In general domains, concepts are often separable.
In medical reports, however, the ``historical comparison'' style is statistically entangled with specific clinical findings~\cite{adila2024steerfair}.
As illustrated in Fig.~\ref{fig:teaser_mesh}, a standard \textbf{Naive Vector (orange)} follows the direction of maximum variance in the difference space.
However, due to entanglement, this vector acts as a diagonal hypotenuse that introduces a lateral shift on the visual manifold. This shift, labeled as the \textbf{Semantic Error (blue)}, corrupts clinical details (e.g., altering ``Pneumonia'' to ``Normal'').
Consequently, standard steering acts as an \textit{indiscriminate} instrument: blindly suppressing it risks removing valid clinical diagnoses along with the historical hallucinations.

To resolve this, we introduce \textbf{Semantically Decoupled Latent Steering (SDLS)}, a framework that prioritizes geometric disentanglement.
Unlike naive approaches that simply subtract the entangled vector, SDLS leverages $QR$ decomposition to strictly enforce orthogonality, thereby geometrically disentangling style from content.
By decomposing the biased direction (i.e., the vector aligned with historical priors), we isolate the \textbf{SDIV (purple)}, a vector strictly orthogonal to the visual manifold.
This geometric constraint ensures the correction removes the bias height (historical style) without introducing the lateral semantic error.
By mathematically enforcing this separation, the $QR$ process acts as a safety filter, constructing a semantic-free intervention vector that targets only the ``historical comparison'' axis.

We position our approach not merely as a correction technique, but as a tool for \textit{mechanistic interpretability}.
By neutralizing the statistical language priors that exacerbate the deviation from the visual manifold (manifesting as the ``Semantic Error'' in Fig.~\ref{fig:teaser_mesh}), SDLS allows the latent trajectory (the sequence of hidden states generated during inference) to be recaptured by the visual manifold. This restores the model's visual grounding, defined as the capacity to rely on image features rather than language priors. This grounding deficit is the root cause of the hallucination~\cite{xiao2025grounding, ding2024image}.
Our contributions are summarized as follows:
\begin{itemize}
    \item We propose SDLS, a training-free steering mechanism that employs $QR$-based orthogonalization to decouple hallucination bias from visual semantics.
    \item  We provide a mechanistic explanation linking prior-comparison hallucinations to visual grounding failures, validated via attention map analysis.
    \item  Utilizing a rigorous Dual-Objective Framework, we demonstrate a ``positive-sum'' outcome on BiomedGPT, reducing hallucination scores by up to 37.3\% on unseen datasets (IU-Xray, CheXpert Plus) without compromising clinical fidelity.
\end{itemize}
Furthermore, our results on LLaVA-Med reveal a critical dependency: effective latent steering requires deep, persistent cross-modal interaction (as in BiomedGPT), whereas decoder-only architectures with visual prefixes may diffuse this information, making it harder to steer linearly.

\section{Related work}
\label{sec:related_work}
\begin{figure*}[t!]
    \centering
    \includegraphics[width=\textwidth]{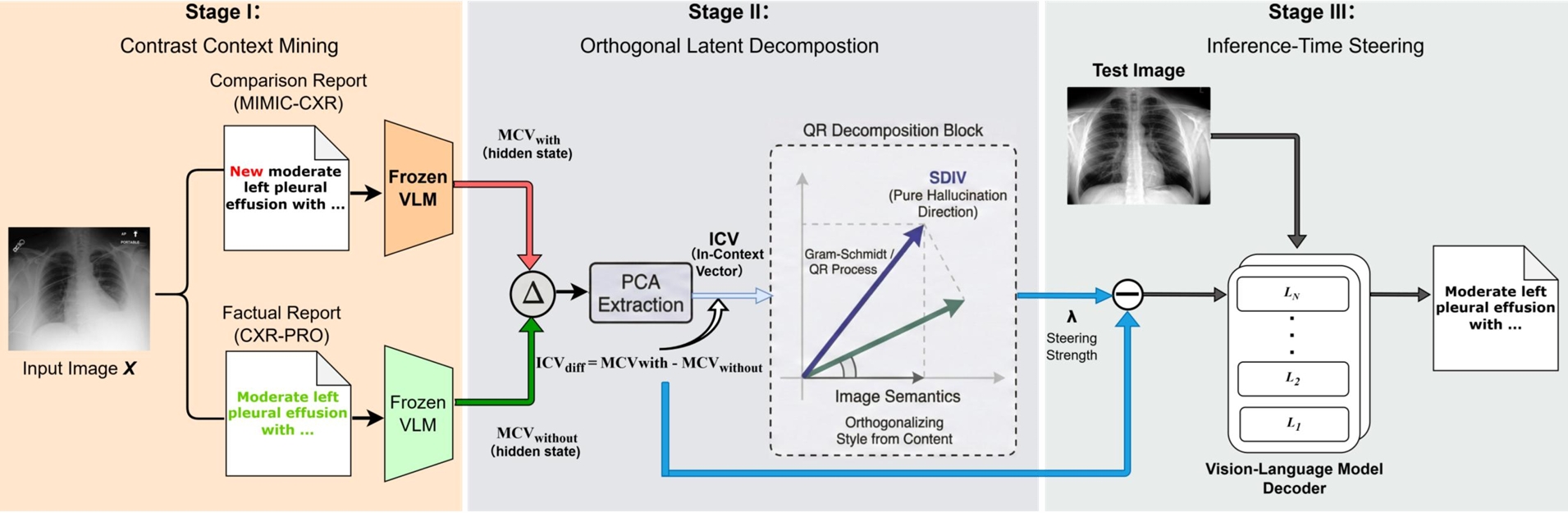} 
    \caption{Schematic overview of the Semantically Decoupled Latent Steering (SDLS) framework. The framework comprises an offline preparation phase (a-b) and an online deployment phase (c). \textbf{(a) Contrastive Context Mining:} Paired reports (hallucinated vs. clean) are encoded to isolate the raw bias representation. \textbf{(b) Orthogonal Latent Decomposition:} A $QR$-based factorization extracts the pure hallucination direction (\sdiv, purple arrow) orthogonal to image semantics. \textbf{(c) Inference-Time Steering:} During test time, the SDIV is injected into decoder layers to suppress prior-comparison tokens without model retraining.}
    \label{fig:framework}
\end{figure*}

\subsection{Radiology Report Generation: Architecture and Evaluation}
\label{sec:rel_rrg}
The field of automated radiology report generation has evolved rapidly from early captioning systems to sophisticated vision-language models (VLMs) \cite{stefanini2022show,ma2025towards}. Early approaches treated report generation as translation, employing CNN-RNN architectures~\cite{jing2017automatic} or joint embedding spaces like TieNet~\cite{wang2018tienet}. The field has since transitioned to Transformer-based encoders~\cite{li2023dynamic,li2021ffa,li2022cross,liu2025hc,li2023auxiliary} and, more recently, general-purpose LLMs adapted for medical tasks~\cite{song2023how,li2024contrastive,lin2023towards}. 
Recent frameworks like Flamingo-CXR~\cite{tanno2025collaboration} and MCVGen~\cite{liu2024mcvgen} further demonstrate the potential of adapting foundation models for zero-shot clinical reporting and human-AI collaboration.
However, ensuring fine-grained alignment between visual features and generated text remains a critical challenge~\cite{liu2024multi,fan2024medical}.

Despite architectural progress, reliable evaluation remains a bottleneck. The standard practice relies on Natural Language Generation (NLG) metrics (e.g., BLEU, ROUGE) and clinical labeling tools like CheXpert~\cite{irvin2019chexpert}. However, recent benchmarks specifically focused on medical hallucinations, such as MedHallTune~\cite{yan2025medhalltune}, demonstrate that these surface-level metrics often fail to capture fine-grained semantic discrepancies and factual errors. Recent comprehensive studies by Yu et al.~\cite{yu2023patterns} demonstrate that surface-level text metrics can diverge significantly from radiologist judgment, particularly for longitudinal findings or subtle disease progressions.
Standard evaluations often aggregate these scores into a single metric, potentially masking specific failure modes. In contrast, our work explicitly decouples evaluation. By employing a Dual-Objective Framework that separates hallucination suppression (measured via prior-checking classifiers) from clinical fidelity (measured via label extraction), we ensure that improvements in one dimension do not silently trade off against the other.

\subsection{The Challenge of Temporal Hallucination in Medical AI}
\label{sec:rel_hallucination}
Hallucination, defined as the generation of non-factual content, is a critical barrier to the deployment of VLMs in healthcare. While general domain hallucinations (e.g., object fabrication) are well-studied~\cite{liu2024survey, guan2024hallusionbench}, medical hallucinations present unique safety risks. Surveys by Zhu et al.~\cite{zhu2025can} and Kim et al.~\cite{kim2025medical} classify \textit{temporal hallucination} (or prior-comparison hallucination) as a high-severity error type. This occurs when a model generates definitive statements about disease progression (e.g., ``no interval change," ``worsening opacity") based solely on a single current image, without access to prior studies.

Ramesh et al.~\cite{ramesh2022filbert} provided a definitive characterization of this issue using the MIMIC-CXR dataset. Their analysis revealed a severe distributional bias: over 76\% of the reports in the training corpus contain references to prior exams. Consequently, models learn a strong statistical prior to generate comparative language as a stylistic default. In a clinical setting, such hallucinations are dangerous; a falsely reported ``stable" condition can delay necessary intervention, while a false ``worsening" can lead to unnecessary procedures~\cite{BelislePipon2024}. This specific crisis of trust necessitates targeted intervention strategies beyond generic quality improvements. Recent studies on adversarial attacks~\cite{Omar2025} further highlight the vulnerability of these models to misleading clinical contexts, reinforcing the need for robust control mechanisms.

\subsection{Mitigation Strategies: Data-Centric vs. Latent Steering}
\label{sec:rel_mitigation}
Current mitigation approaches fall into two paradigms. Unlike model editing techniques such as ROME~\cite{meng2022locating}, which permanently modify parameters to update factual knowledge, our work focuses on inference-time control.

\subsubsection{Data-Centric Interventions (Retraining)}
The prevailing strategy addresses the problem at the data source via filtering or rewriting contaminated samples (e.g., CXR-PRO) followed by retraining~\cite{ramesh2022filbert}.
While this approach effectively reduces prior-comparison errors, it incurs significant operational costs. Retraining large foundation models is computationally expensive and must be repeated for every model update~\cite{pham2024towards}. More critically, this method effectively permanently impairs the model. By removing all valid comparisons from the training distribution, the model permanently loses the capability to perform longitudinal analysis, even when valid historical context is provided. It treats the symptom (dataset bias) but impairs the model's general utility.

\subsubsection{Inference-Time Representation Control}
Our work advocates for a model-centric, inference-time approach, building on the principles of Representation Engineering (RepE)~\cite{zou2023representation}. This paradigm posits that high-level semantic concepts are encoded as linear directions in the model's activation space, a hypothesis recently supported by theoretical analyses of LLM latent structures~\cite{saglam2025large, siddique2025shifting}. Techniques such as \textit{Activation Addition}~\cite{turner2023actadd} and \textit{Function Vectors}~\cite{todd2023function} have demonstrated that adding ``steering vectors" to hidden states can modulate attributes like sentiment, honesty, or refusal behavior without parameter optimization. Alternative inference-time strategies, such as Visual Contrastive Decoding (VCD)~\cite{leng2024mitigating}, attempt to mitigate hallucinations by contrasting output distributions from perturbed inputs. Unlike contrastive decoding strategies that double the inference cost, SDLS introduces negligible latency as it operates within a single forward pass. Building on the foundational concept of activation addition~\cite{turner2023activation}, Liu et al.~\cite{liu2024icv} further formalized this with \textit{In-Context Vectors (ICV)}, extracting steering directions from contrastive few-shot demonstrations\cite{liu2025reducing}.

\subsection{Mechanistic Underpinnings: The Modality Gap}
\label{sec:rel_mechanism}
Finally, our intervention is grounded in the mechanistic theory of multimodal generation. Research in image captioning has identified a tension between \textit{Language Priors} and \textit{Visual Grounding} \cite{qiao2021visual}. Studies by Ding et al.~\cite{ding2024image} and Xiao et al.~\cite{xiao2025grounding} argue that hallucinations arise when the decoder over-relies on linguistic co-occurrence statistics, effectively ignoring the visual encoder. This phenomenon is exacerbated as language models become stronger and contexts grow longer~\cite{song2023how}.

Building on these insights, we posit that prior-comparison hallucination acts as a specific clinical manifestation of this grounding deficit.
Our mechanistic analysis reveals that \textit{one of the mechanisms} driving this phenomenon is the failure of visual grounding~\cite{xiao2024visual}, where the model's focus drifts from specific visual features to generic linguistic patterns of comparison.
Therefore, rather than claiming to fully bridge the modality gap, our intervention targets this specific grounding instability, aiming to steer latent representations towards better alignment with the visual context.

\section{Method}
\label{sec:method}
This section details the comprehensive framework developed to suppress and analyze prior-comparison hallucinations in radiology report generation. Our approach consists of three main phases, as illustrated in Fig.~\ref{fig:framework}.
(a) Contrastive Context Mining: To isolate the representation of hallucination bias, we encode contrastive pairs of reports.
(b) Orthogonal Latent Decomposition: We extract a purified intervention vector via $QR$-based factorization.
(c) Inference-Time Steering: The derived vector is injected to suppress prior comparisons dynamically.

\subsection{Task and Scope}
\label{ssec:task_scope}
We restrict both generation and evaluation to the \emph{Impression} section of radiology reports.
Given a frozen backbone and a reference \emph{Impression}, the objective is to suppress historical or non-current statements in generated \emph{Impressions} without retraining.


\subsection{SDIV Formulation via $QR$-based Semantic Decomposition}
\label{ssec:hciv_formulation_qr}
Our intervention targets representative VLMs for chest X-ray report generation, covering both Encoder-Decoder and Decoder-only architectures (detailed in Section~\ref{sec:setup}).
In all cases, we operate only at inference time and leave model weights and training data unchanged.
For each backbone, we estimate the SDIV from paired reports that differ only in their use of historical information.
These pairs are identified across multiple large-scale chest X-ray datasets (including MIMIC-CXR~\cite{johnson2019mimic_cxr}, IU-Xray~\cite{demner2016preparing}, and CheXpert Plus~\cite{chambon2024chexpert}) using a standardized cleaning pipeline.

\subsubsection{Notation and Representation Basis}
For a given backbone $m$, input image $x$, and a target text sequence $r$ (either $r^{\mathrm{hist}}$ or $r^{\mathrm{curr}}$), we perform a forward pass through the model to extract its internal representations. Let $\mathbf{h}^{(m)}_{\ell}(x,r) \in \mathbb{R}^{d_m}$ denote the hidden state of the \textbf{final token} of the sequence $r$ at the output of decoder layer $\ell$. To provide a comprehensive representation, we consider the word embedding layer as the initial layer (layer 0) in the decoder stack. The multi-layer contextual vector (MCV), denoted as $\mathbf{z}^{(m)}(x,r)$, is then defined as the concatenation of these hidden states across all $L_m$ selected layers:
\begin{equation}
    \mathbf{z}^{(m)}(x,r) = [\mathbf{h}^{(m)}_{0}(x,r); \mathbf{h}^{(m)}_{1}(x,r); \dots; \mathbf{h}^{(m)}_{L_m-1}(x,r)] \in \mathbb{R}^{D_m},
\end{equation}
where $D_m = L_m d_m$. For each paired sample $(r^{\mathrm{hist}}, r^{\mathrm{curr}})$ linked to the same image, we form a difference vector which captures the semantic shift associated with the presence of historical language:
\begin{equation}
    \boldsymbol{\delta}^{(m)} = \mathbf{z}^{(m)}(x, r^{\mathrm{hist}}) - \mathbf{z}^{(m)}(x, r^{\mathrm{curr}}).
\end{equation}
Stacking all such difference vectors for backbone $m$ yields a matrix
$\mathbf{D}^{(m)} = [\boldsymbol{\delta}^{(m)}_1, \dots, \boldsymbol{\delta}^{(m)}_{N_m}]$.
This aggregation allows us to analyze the statistical structure of the bias, hypothesizing that while individual vectors contain noise, the collective distribution reveals the entangled direction of historical style.
This matrix forms the basis for constructing our various intervention vectors.
\begin{algorithm}[t]
\caption{Semantically Decoupled Latent Steering (SDLS)}
\label{alg:sdls}
\small
\textbf{Input:} Pre-trained VLM $\mathcal{M}$, Image $x$, Steering Strength $\lambda$, Contrastive Pairs $\mathcal{P} = \{(r^{\mathrm{hist}}_i, r^{\mathrm{curr}}_i)\}_{i=1}^N$ \\
\textbf{Output:} Generated Report $y$
\begin{algorithmic}[1]
\STATE \textcolor{blue}{\textit{// Phase 1: SDIV Construction (Offline)}}
\STATE Collect difference vectors $\mathbf{D} \leftarrow \{\mathbf{z}(\text{hist}) - \mathbf{z}(\text{curr}) \mid \forall \text{pair} \in \mathcal{P}\}$
\STATE Decompose $\mathbf{D}$ into semantic classes $C_1, \dots, C_k$ via LLM
\FOR{each class $k$}
    \STATE $\mathbf{v}_k \leftarrow \text{PCA}(\mathbf{D}_{C_k}, \text{top-1})$ \COMMENT{Extract dominant direction per class}
\ENDFOR
\STATE $\mathbf{Q}, \mathbf{R} \leftarrow \text{QR\_Decomposition}([\mathbf{v}_1, \dots, \mathbf{v}_k])$
\STATE $\mathbf{v}_{\text{sdiv}} \leftarrow \text{Normalize}(\text{Mean}(\text{Columns}(\mathbf{Q})))$ \COMMENT{Compute Consensus Direction}

\STATE \textcolor{blue}{\textit{// Phase 2: Inference Steering (Online)}}
\FOR{step $t = 1, \dots, T$}
    \STATE $\mathbf{h}_t \leftarrow \mathcal{M}_{\text{encode}}(x, y_{<t})$
    \STATE $\tilde{\mathbf{h}}_t \leftarrow \mathbf{h}_t + \lambda \cdot \mathbf{v}_{\text{sdiv}}$ \COMMENT{Inject steering vector}
    \STATE $y_t \sim \text{Softmax}(\mathcal{M}_{\text{head}}(\tilde{\mathbf{h}}_t))$
\ENDFOR
\end{algorithmic}
\end{algorithm}

\subsubsection{Baseline ICV Formulations for Ablation}
\label{ssec:baseline_icvs}
To establish strong baselines and diagnose model behavior, we formulate two types of ICV based on principal component analysis (PCA) of the difference matrix $\mathbf{D}^{(m)}$.
\paragraph{Global ICV (Broad Distributional Bias)}
This vector captures the average semantic direction of ``history'' across the entire dataset. We compute the empirical mean of all difference vectors, $\boldsymbol{\mu}^{(m)}$, and project it onto the top-$k$ principal components of the centered difference matrix $\mathbf{D}^{(m)}$ to isolate the principal directions of variation and suppress random noise:
\begin{equation}
    \mathbf{v}^{(m)}_{\text{icv-global}}(k) = \mathbf{U}^{(m)}_k (\mathbf{U}^{(m)}_k)^{\top} \boldsymbol{\mu}^{(m)},
\end{equation}
where $\mathbf{U}^{(m)}_k$ contains the top-$k$ principal components. 
This vector captures the macro-level statistical prior, representing the broad tendency of the model to drift into historical narration due to dataset imbalance. However, this comes at the cost of potential noise and semantic entanglement.

\paragraph{Curated-Specific-50 ICV (Surgical Removal)}
This high-precision vector is derived from a manually curated subset of 50 report pairs characterized by minimal, single-word edits to maximize semantic isolation, ensuring the difference vector captures only the presence of history without confounding syntactic shifts. Using only these pairs, we compute a specialized mean difference $\boldsymbol{\mu}^{(m)}_{\mathrm{S50}}$ and its principal components $\mathbf{U}^{(m)}_{k,\mathrm{S50}}$ to define:
\begin{equation}
    \mathbf{v}^{(m)}_{\text{icv-s50}}(k) = \mathbf{U}^{(m)}_{k,\mathrm{S50}} (\mathbf{U}^{(m)}_{k,\mathrm{S50}})^{\top} \boldsymbol{\mu}^{(m)}_{\mathrm{S50}}.
\end{equation}
In contrast to the Global ICV, this vector targets the ``surgical removal'' of specific, high-frequency hallucination triggers (e.g., ``unchanged'', ``stable''). It acts as a micro-level intervention, offering high precision but lower coverage of the diverse historical expressions found in the wild.


\subsubsection{Default SDIV via Automated Semantic Disentanglement}
To create a more stable and disentangled intervention direction, our default SDIV builds upon the concept of semantic ICV. To achieve this, we employ a LLM, (\texttt{gpt-oss:20b}) to automatically parse and disentangle the complex narratives within the ``with-history" reports. For each report, the LLM deconstructs it into a list of individual clinical findings and assigns each finding to one of $C$ pre-defined temporal semantic classes (e.g., improved, worsened, device-related, unchanged). While clinical semantics vary across these categories, the historical style remains constant. Averaging their directions allows the variable clinical content to cancel out via destructive interference. This process isolates the invariant stylistic bias.

This automated process yields a set of semantically categorized finding pairs. For each semantic class $c$ and backbone $m$, we then form a matrix $\mathbf{D}^{(m)}_c$ of difference vectors. We compute the first principal component $\mathbf{p}^{(c)}_m$ of this matrix, which represents a specific semantic ICV. We stack these $C$ semantic directions into a matrix $\mathbf{V}^{(m)} = [\mathbf{p}^{(1)}_m, \dots, \mathbf{p}^{(C)}_m]$.

\subsubsection{From PCA Identification to $QR$ Orthogonalization}
While PCA identifies axes of maximum variance, it assumes that the dominant direction is purely the target bias. However, in the radiology latent space, the ``historical style'' direction is often statistically entangled with clinical semantics (e.g., specific disease progression patterns). Consequently, the principal component of the raw difference matrix may inadvertently capture clinical content common to the dominant class (e.g., maximizing variance by aligning with the prevalent ``stability'' concept rather than the pure historical style).

To address this, we employ $QR$ decomposition to stabilize the basis of our extracted semantic directions. Let $\mathbf{V}^{(m)} = [\mathbf{p}^{(1)}_m, \dots, \mathbf{p}^{(C)}_m]$ be the matrix where each column represents the primary difference vector for a distinct temporal semantic category (e.g., \textit{worsened} vs. \textit{improved}). We posit that these vectors share a common component, specifically the ``historical reporting style,'' while varying in their clinical specifics.

We perform a $QR$ factorization $\mathbf{V}^{(m)} = \mathbf{Q}^{(m)} \mathbf{R}^{(m)}$, where the columns of $\mathbf{Q}^{(m)} = [\mathbf{q}^{(1)}_m, \dots, \mathbf{q}^{(C)}_m]$ form an orthonormal basis spanning the subspace of the observed differences. Unlike PCA, which seeks variance, this step normalizes the contribution of each semantic category. 
Crucially, since the clinical semantics (e.g., pathology changes) vary across the categories $c=1 \dots C$ while the historical style remains constant, the \textit{mean} of these basis vectors serves to construct a consensus direction. Because the historical style is invariant across categories while clinical content varies, this operation constructively interferes with the shared stylistic component while marginalizing the category-specific clinical variations. The final SDIV is defined as the normalized consensus direction:
\begin{equation}
    \mathbf{v}^{(m)}_{\text{sdiv}} = \mathrm{norm}\!\left( \frac{1}{C} \sum_{c=1}^{C} \mathbf{q}^{(c)}_m \right).
\end{equation}
This vector represents the semantic-free ``noise'' of historical style, effectively decoupled from the specific clinical signals inherent in the source pairs. Note that $\mathbf{v}^{(m)}_{\text{sdiv}}$ has the same dimensionality $D_m$ as the target hidden states, enabling direct additive intervention.

\subsection{Inference-Time Intervention Strategies}
\label{ssec:intervention_strategies_method}
We apply the constructed vectors to the model's hidden states at inference time.
To systematically probe the model's behavior and identify the optimal control mechanism, we define a spectrum of injection strategies targeting different depths and modules of the architecture.

\subsubsection{Norm-Preserving Additive Steering}
For most strategies, we employ a norm-preserving update to prevent distribution shift.
Unconstrained addition of steering vectors can inadvertently shift the feature magnitude, pushing representations out of the distribution expected by subsequent layers (e.g., LayerNorm).
The updated hidden state $\tilde{\mathbf{h}}$ is computed to alter the representation's direction while strictly maintaining its magnitude:
\begin{equation}
    \tilde{\mathbf{h}} = \|\mathbf{h}\|_2 \,\mathrm{norm}\!\left(\frac{\mathbf{h}}{\|\mathbf{h}\|_2} + \lambda \mathbf{v}^{(m)}_{\ell}\right),
    \label{eq:norm_preserving_injection}
\end{equation}
where $\mathrm{norm}(\cdot)$ denotes L2 normalization and $\lambda$ is the steering strength(typically $\lambda < 0$ for suppression). Since the steering involves only element-wise addition to the hidden states, it incurs negligible computational overhead compared to the standard autoregressive generation process. Details of the complete inference procedure are summarized in Algorithm~\ref{alg:sdls}.


\subsubsection{Strategic Injection Loci}
We instantiate this steering mechanism via four distinct strategies, ranging from global to targeted interventions:

\paragraph{Global Injection}
This baseline strategy injects the intervention vector into the outputs of \textbf{all} decoder layers, including the word embedding layer. It aims for a persistent, global steering effect throughout the entire generation process, simulating a fundamental shift in the model's operating mode.

\paragraph{SteerFair (Targeted Attention Steering)}
Acknowledging that hallucinations often stem from specific modular failures, this strategy targets the transformer's internal sub-layers.
\begin{itemize}
    \item \textbf{LayerOutput:} Injects the vector at the final output of each transformer block (post-FeedForward).
    \item \textbf{AttentionOutput:} Injects the vector specifically at the output of the self-attention mechanism. This variant tests the hypothesis that historical priors are primarily encoded in the attention mixing weights.
\end{itemize}

\paragraph{GentleInject (Early Perturbation)}
This strategy applies a strong, non-norm-preserving additive shift only to the start-of-sequence ([CLS] or equivalent) token at the \textbf{first decoder layer}. It tests whether a single, early perturbation is sufficient to set the trajectory for the entire generation, minimizing computational overhead.

\paragraph{ICV-Token (Input-Level Prompting)}
Instead of modifying hidden states, this strategy treats the intervention vector as a pseudo-token. The vector is projected to the model's hidden dimension and prepended to the decoder's input sequence, influencing generation via the standard causal attention mechanism.

\subsection{Comprehensive Evaluation Protocol: Metrics and Objectives}
\label{ssec:comprehensive_eval}

To strictly assess the "positive-sum" hypothesis, we employ a dual-objective framework that simultaneously measures hallucination suppression and clinical fidelity.

\subsubsection{Suppression Metric: History-comparison Span Rate (HSR)}
To quantify the presence of unwanted historical semantics, we define the HSR.
Given a generated report $y$, we identify all occurrences of cue words from a curated dictionary, extended from the standard lexicon defined in FilBERT~\cite{ramesh2022filbert} to include common historical expressions, detailed in Table~\ref{tab:hsr_triggers}.
These cues are grouped into semantic categories such as stability, comparison, and progression.
Using a regex-based search, we calculate the \textbf{Token-level HSR}, defined as the fraction of tokens in the report that are part of any cue span:
\begin{equation}
    \text{HSR}(y) = \frac{|S(y)|}{|y|},
\end{equation}
where $S(y)$ is the set of tokens within matched cue spans.

\subsubsection{Dual-Objective Objectives}
\label{ssec:dual_objectives}
We formulate the evaluation as a constrained optimization problem. Let $\mathcal{D}$ be the evaluation set and $b$ the baseline.
For method $m$, we define:

\paragraph{1. Suppression Objective}
We aim to maximize the reduction in historical artifacts. The primary metric is the improvement in HSR relative to the baseline:
\begin{equation}
    \Delta\mathrm{HSR}(m) = \overline{\mathrm{HSR}}(b)-\overline{\mathrm{HSR}}(m).
\end{equation}
We also report reductions in the probabilities assigned by classifier judges (FilBERT~\cite{ramesh2022filbert} and BERT), denoted as $\Delta p^{\mathrm{fil}}$ and $\Delta p^{\mathrm{bert}}$.

\paragraph{2. Fidelity Objective}
We must maintain clinical accuracy. We measure this using $\mathrm{F1}^{\mathrm{macro}}(m)$ derived from CheXpert labeler~\cite{irvin2019chexpert} on the generated Impressions.

\paragraph{Selection Rule}
A method $m$ is considered a valid improvement if and only if it improves suppression without degrading fidelity:
\begin{equation}
\mathrm{F1}^{\mathrm{macro}}(m)\ge \mathrm{F1}^{\mathrm{macro}}(b)\quad \text{and}\quad \Delta\mathrm{HSR}(m)>0.
\end{equation}

\begin{table}[h!]
\centering
\caption{Dictionary of Cue Words and Phrases Used for HSR Calculation, Grouped by Semantic Category.}
\label{tab:hsr_triggers}
\renewcommand{\arraystretch}{1.2}
\begin{tabularx}{\columnwidth}{l X}
\hline
\hline
\textbf{Category} & \textbf{Cue Phrases} \\
\hline
Stability & stable, unchanged, no change, no significant change, no interval change, persistent, remains, similar, chronic, long-standing \\
Comparison & prior, previous, compared with, compared to, since the previous, again seen \\
Progression & worsened, increased, larger, more, progression, development of, now demonstrates \\
Improvement & improved, decreased, smaller, less, resolved, resolution, clearing \\
\hline
\multicolumn{2}{p{\dimexpr\columnwidth-2\tabcolsep\relax}}{\footnotesize{\textit{Note:} A separate list of negative phrases (e.g., 'no prior') is used for exclusion during analysis.}} \\
\hline
\hline
\end{tabularx}
\end{table}

\subsection{Mechanistic Probe I: $\Delta$logit Suppression Curve}
\label{ssec:mech_probe_logit}
To directly probe the intervention's impact on the model's next-token predictive distribution, we measure the change in the logits assigned to historical cue tokens. For each cue token $w$ from our dictionary $\mathcal{C}$ (Table~\ref{tab:hsr_triggers}), we compute the change in its logit value as a function of the intervention strength $\lambda$:
\begin{equation}
    \Delta\text{logit}(w, \lambda) = \mathbb{E}_{t, x} \left[ z_t^{(\lambda)}(w, x) - z_t^{(0)}(w, x) \right],
\end{equation}
where $z_t^{(\lambda)}(w, x)$ is the logit for token $w$ given input $x$ and context at step $t$ under intervention strength $\lambda$. A negative $\Delta\text{logit}$ indicates direct suppression. We analyze the resulting $\Delta\text{logit}$ vs. $|\lambda|$ curve to verify a dose-dependent effect. We use the absolute value $|\lambda|$ to represent the increasing intervention magnitude regardless of the vector's sign, facilitating the visualization of a dose-dependent suppression effect.

\subsection{Mechanistic Probe II: Cross-Attention Analysis}
\label{ssec:mech_probe_attention}
To understand the underlying cause of prior-comparison hallucinations, we conduct a qualitative analysis of the decoder's cross-attention mechanism. This probe aims to determine whether these hallucinations are primarily language-model artifacts or are erroneously grounded in visual features. During generation of key tokens, we visualize the cross-attention weights, $A^{\text{cross}}$, which map decoder tokens to encoder (image patch) tokens. By comparing the attention patterns for factual, image-grounded terms versus hallucinated historical terms, we can identify potential failures in the model's visual grounding process. The empirical results of this analysis, detailed in Section~\ref{ssec:results_mechanism}, showing that prior-comparison tokens systematically fail to attend to relevant visual regions compared to factual findings.

\subsection{Clinical-Semantics Metrics on Impression}
\label{ssec:quality_metrics}
We assess clinical semantics with the Stanford CheXpert labeler~\cite{irvin2019chexpert} (enhanced with NegBio dependency parsing~\cite{peng2018negbio}) applied to generated \emph{Impressions}. The uncertainty policy is set to \texttt{zero}. 
To evaluate clinical fidelity, we report the \textbf{Macro-F1} (arithmetic mean of F1 scores across the 14 CheXpert labels) and \textbf{Micro-F1} (computed globally across all samples and labels). 
Standard NLG metrics (BERTScore~\cite{zhang2020bertscore}, ROUGE~\cite{lin2004rouge}, METEOR~\cite{banerjee2005meteor}, BLEU~\cite{Papineni2002BleuAM}) are also reported for completeness.

\section{Experimental Setup}
\label{sec:setup}
All experiments are conducted under a unified protocol to evaluate inference-time latent space interventions for controlling prior-comparison hallucinations in radiology reports. The framework employs a paired-control design: for each input case, a baseline report is generated alongside reports from multiple intervention conditions, enabling direct, case-matched comparisons. All generations and evaluations are restricted to the \emph{Impression} section. The entire process is executed at inference time with no modification to the underlying model's weights.
\begin{table*}[h!]
    \centering
    \caption{\textbf{Metrics Overview Figures.} This table details the specific intervention method and configuration. Note that configurations may vary slightly across different backbone models.}
    \label{tab:overview_legend}
    \renewcommand{\arraystretch}{1.2}
    \begin{tabular}{c l l l}
        \toprule
        \textbf{Position} & \textbf{Configuration on BiomedGPT} & \textbf{Configuration on IAMJB (VED)} & \textbf{Configuration on LLaVA-Med} \\
        \midrule
        B & Baseline & Baseline & Baseline \\
        \midrule
        \multicolumn{4}{l}{\textit{--- GentleInject Methods (Varying Strength $\lambda$) ---}} \\
        1 & GentleInject ($\lambda=-25.0$) & GentleInject ($\lambda=-25.0$) & GentleInject ($\lambda=-25.0$) \\
        2 & GentleInject ($\lambda=-20.0$) & GentleInject ($\lambda=-20.0$) & GentleInject ($\lambda=-20.0$) \\
        3 & GentleInject ($\lambda=-15.0$) & GentleInject ($\lambda=-15.0$) & GentleInject ($\lambda=-15.0$) \\
        4 & GentleInject ($\lambda=-10.0$) & GentleInject ($\lambda=-10.0$) & GentleInject ($\lambda=-10.0$) \\
        5 & GentleInject ($\lambda=-5.0$) & GentleInject ($\lambda=-5.0$) & GentleInject ($\lambda=-5.0$) \\
        \midrule
        \multicolumn{4}{l}{\textit{--- Fine-grained Methods (Grouped by Strength $\lambda$) ---}} \\
        6 & SteerFair\_LayerOutput ($\lambda=-0.5$) & SteerFair\_AttentionOutput ($\lambda=-0.5$) & SteerFair\_AttentionOutput ($\lambda=-0.5$) \\
        7 & Global Injection ($\lambda=-0.5$) & SteerFair\_LayerOutput ($\lambda=-0.5$) & SteerFair\_LayerOutput ($\lambda=-0.5$) \\
        8 & SteerFair\_AttentionOutput ($\lambda=-0.5$) & Global Injection ($\lambda=-0.5$) & Global Injection ($\lambda=-0.5$) \\
        \addlinespace
        9 & SteerFair\_AttentionOutput ($\lambda=-0.4$) & Global Injection ($\lambda=-0.4$) & SteerFair\_AttentionOutput ($\lambda=-0.4$) \\
        10 & Global Injection ($\lambda=-0.4$) & SteerFair\_LayerOutput ($\lambda=-0.4$) & Global Injection ($\lambda=-0.4$) \\
        11 & SteerFair\_LayerOutput ($\lambda=-0.4$) & SteerFair\_AttentionOutput ($\lambda=-0.4$) & SteerFair\_LayerOutput ($\lambda=-0.4$) \\
        \addlinespace
        12 & Global Injection ($\lambda=-0.3$) & Global Injection ($\lambda=-0.3$) & SteerFair\_AttentionOutput ($\lambda=-0.3$) \\
        13 & SteerFair\_AttentionOutput ($\lambda=-0.3$) & SteerFair\_LayerOutput ($\lambda=-0.3$) & Global Injection ($\lambda=-0.3$) \\
        14 & SteerFair\_LayerOutput ($\lambda=-0.3$) & SteerFair\_AttentionOutput ($\lambda=-0.3$) & SteerFair\_LayerOutput ($\lambda=-0.3$) \\
        \addlinespace
        15 & SteerFair\_LayerOutput ($\lambda=-0.2$) & Global Injection ($\lambda=-0.2$) & SteerFair\_LayerOutput ($\lambda=-0.2$) \\
        16 & SteerFair\_AttentionOutput ($\lambda=-0.2$) & SteerFair\_LayerOutput ($\lambda=-0.2$) & SteerFair\_AttentionOutput ($\lambda=-0.2$) \\
        17 & Global Injection ($\lambda=-0.2$) & SteerFair\_AttentionOutput ($\lambda=-0.2$) & Global Injection ($\lambda=-0.2$) \\
        \addlinespace
        18 & SteerFair\_AttentionOutput ($\lambda=-0.1$) & SteerFair\_AttentionOutput ($\lambda=-0.1$) & Global Injection ($\lambda=-0.1$) \\
        19 & Global Injection ($\lambda=-0.1$) & Global Injection ($\lambda=-0.1$) & SteerFair\_LayerOutput ($\lambda=-0.1$) \\
        20 & SteerFair\_LayerOutput ($\lambda=-0.1$) & SteerFair\_LayerOutput ($\lambda=-0.1$) & SteerFair\_AttentionOutput ($\lambda=-0.1$) \\
        \midrule
        \multicolumn{4}{l}{\textit{--- Input-Level Methods ---}} \\
        21 & ICV\_Token\_Method & ICV\_Token\_Method & ICV\_Token\_Method \\
        22 & --- & Encoder\_Concat\_Method & Encoder\_Concat\_Method \\
        \bottomrule
    \end{tabular}
\end{table*}
\subsection{Datasets and Benchmarks}
\label{ssec:data_curation_exp}
We validate our framework on three diverse chest X-ray benchmarks to ensure robustness and generalization.
\paragraph{MIMIC-CXR-JPG}
We use the MIMIC-CXR-JPG v2.0.0 dataset~\cite{johnson2019mimic_cxr} as our primary testbed.
Adopting the standard splits, we define ``with-history'' reports as the original texts and ``no-history'' reports as the cleaned counterparts from CXR-PRO~\cite{ramesh2022filbert}.
Our rule-based pipeline aligns these to yield \Npairs\ paired reports with textual differences for vector construction.
\paragraph{IU-Xray and CheXpert Plus}
To verify the method's transferability, we further include the IU-Xray~\cite{demner2016preparing} and the recently released CheXpert Plus~\cite{chambon2024chexpert} datasets.
For all evaluations, we use a fixed random sub-sample of \Ntest\ test cases from the respective test sets to maintain computational feasibility, given the high cost of the LLM-based metric evaluations.

\subsection{Backbone Vision-Language Models}
\label{sec:backbone_vlms_experiment}
We evaluate our methodology on three VLMs, selected to represent distinct architectural paradigms crucial for understanding the intervention's boundaries.

\paragraph*{BiomedGPT-Base}
Our primary evaluation platform is BiomedGPT-Base, an advanced encoder--decoder model based on the One-For-All (OFA) architecture which features multiple cross-attention blocks. We treat each of its decoder blocks as a layer. For interventions, we register norm-preserving additive hooks on grouped decoder layers (an early, a middle, and a late group). Its state-of-the-art architecture makes it the primary testbed for SDIV efficacy.

\paragraph*{Vision-Encoder-Decoder (VED) model}
We use the \url{IAMJB/chexpert-mimic-cxr-impression-baseline} model, hereafter referred to as the VED model, which couples a SwinV2 encoder with a two-layer BertGeneration decoder and explicit cross-attention. For this model, we define three canonical injection sites: the output of the decoder word embedding layer (Site 1), the first decoder transformer layer (Site 2), and the final decoder layer (Site 3). Its simpler structure provides a valuable setting for mechanistic analysis and for testing sensitivity to data purity.

\paragraph*{LLaVA-Med v1.5}
As a structural contrast, we include LLaVA-Med v1.5. This model projects image features into the language embedding space and prepends them as special tokens to a LLaMA-style decoder, relying on self-attention for vision-language interaction post-projection.
The behavior observed in LLaVA-Med allows us to investigate the necessity of deep cross-modal interaction for linear steering. By comparing this visual-prefix architecture with cross-attention-based models, we aim to isolate the architectural prerequisites for effective latent space intervention.

\subsection{Intervention Vector Arsenal}
\label{ssec:vector_arsenal}
For each backbone model, we construct the suite of intervention vectors detailed in Section~\ref{ssec:hciv_formulation_qr} from the 5,000 curated report pairs. This includes our main proposal, the \textbf{SDIV}, and two ablative baselines: the \textbf{Global ICV} and the \textbf{Curated-Specific-50 ICV}. For the ICV-based vectors, a key hyperparameter is the number of principal components, $k$; we sweep $k \in \{1, 2, 3, 5, 10, 30, 100\}$ and report the results from the best performing configuration.

\subsection{Inference-Time Intervention Strategies}
\label{ssec:intervention_strategies}
The constructed vectors are deployed using a variety of strategies that differ in their injection site, scope, and application logic. These strategies, summarized in Table~\ref{tab:intervention_families}, allow us to probe the model's internal mechanisms. The primary mechanism for most strategies is the norm-preserving additive injection defined in Equation~\ref{eq:norm_preserving_injection}, designed to alter a hidden state's direction while minimizing disruption to its magnitude.

\begin{table}[h!]
\centering
\caption{Intervention Strategies and Their Mechanisms.}
\label{tab:intervention_families}
\renewcommand{\arraystretch}{1.3}
\begin{tabularx}{\columnwidth}{l X}
\hline
\hline
\textbf{Strategy} & \textbf{Mechanism and Implementation Detail} \\
\hline
\texttt{Global Injection} & Injects the vector into the outputs of \textbf{all} decoder layers, including the word embedding layer. Aims for a global, persistent steering effect throughout the entire decoding stack. \\
\texttt{SteerFair} & A more targeted approach with two variants: \texttt{LayerOutput} injects into the final output of each transformer layer (excluding embeddings), while \texttt{AttentionOutput} injects deeper, specifically at the self-attention module's output. \\
\texttt{GentleInject} & Applies a strong, non-norm-preserving additive shift (\verb|hs[:, 0, :] += lam * vec|) only to the [CLS] token's representation at the first decoder layer, testing the impact of an early, high-impact perturbation. \\
\texttt{ICV-Token} & Treats the vector as a pseudo-token. It is projected to the model's hidden dimension and prepended to the decoder's input sequence, influencing generation from the initial context. \\
\texttt{Encoder-Concat} & The vector is projected and concatenated to the visual encoder's output sequence, directly modifying the visual context available to the decoder. This method was found to be incompatible with BiomedGPT's \texttt{.generate} function and was thus excluded for that backend. \\
\hline
\hline
\end{tabularx}
\end{table}

\subsection{Robustness and Ablation Setup}
The specificity of the SDIV effect is tested via a suite of controls and ablations.
Negative controls include interventions with random vectors, vectors with shuffled components (preserving magnitude but destroying direction), and vectors orthogonal to the main SDIV direction.
Furthermore, we conduct a style orthogonalization study to ensure our intervention targets historical semantics rather than mere writing style.
We construct a ``style subspace" using the principal components of the hidden state representations from a large corpus of history-free reports. We then project our Global ICV onto this subspace and subtract the projection, yielding a ``style-orthogonalized'' ICV, a technique inspired by iterative nullspace projection~\cite{ravfogel2020null}.
This orthogonalized vector is used in a controlled experiment to verify that the core suppression effect is preserved. Ablation studies also compare the efficacy of different intervention strategies (e.g., all-layer vs. targeted-layer injection) as detailed in our experimental setup.

\subsection{Clinical-Semantics Metrics Configuration}
We assess clinical fidelity using the Stanford CheXpert labeler~\cite{irvin2019chexpert} enhanced with NegBio~\cite{peng2018negbio}.
The uncertainty policy is set to \texttt{zero}. 
We report \textbf{Macro-F1} (mean across 14 labels) and \textbf{Micro-F1} (global calculation).
Standard NLG metrics (BERTScore, ROUGE, METEOR, BLEU) are also computed for completeness.

\subsection{Statistical Analysis}
All primary results are reported with 95\% confidence intervals (CIs) estimated via non-parametric bootstrapping (10,000 resamples). Paired tests are used to account for the within-subject nature of the interventions, controlling the false discovery rate at $\alpha=0.05$ (Benjamin-Hochberg).

\subsection{Experimental Protocol}
Our main experiment consists of a large-scale evaluation grid, crossing the three backbone models with the three vector types and the various intervention strategies. The intervention strength, $\lambda$, is swept across a range tailored to each strategy's sensitivity. For the fine-grained \texttt{Global Injection} and \texttt{SteerFair} variants, we use a grid of smaller values (e.g., focusing on the suppression range $\lambda \in \{-0.1, \dots, -0.5\}$). For the high-impact \texttt{GentleInject} strategy, a grid of larger values is used (e.g., $\lambda \in \{-5, \dots, -25\}$).

\paragraph{For BiomedGPT and VED models}
Interventions are applied by registering forward hooks on the target PyTorch modules. Generation is performed using the standard \texttt{.generate()} method with beam search (4 beams, \texttt{no\_repeat\_ngram\_size=3} for BiomedGPT, \texttt{repetition\_penalty=1.1} for VED).

\paragraph{For LLaVA-Med}
Due to its architecture, a standard hook-based approach with \texttt{.generate()} is not feasible. We implement a custom auto-regressive decoding loop to enable per-token intervention. At each generation step, a forward pass is performed to get the logits for the next token. Before the next token is chosen, our intervention hooks modify the hidden states in the specified layers. The strength of the injection can be dynamically decayed across generation steps, as is the case for some strategies (e.g., \texttt{GentleInject}).

\subsection{Evaluation Metrics: Semantics and Decoupling}
\label{ssec:exp_evaluation_metrics}

\subsubsection{Semantic Evaluation with Calibrated Classifier Judges}
We use two classifiers, FilBERT~\cite{ramesh2022filbert} and a custom-trained BERT, as automated judges to provide a sentence-level semantic evaluation.
For a report $y$, each judge outputs a probability $P(\text{no history}|y)$.
The reliability of these judges is rigorously assessed on a held-out set using:
\begin{itemize}
    \item \textbf{Discrimination:} Area Under the Receiver Operating Characteristic curve (AUROC) and the Precision-Recall curve (AUPRC).
    \item \textbf{Calibration:} Expected Calibration Error (ECE)~\cite{guo2017calibration} to measure the consistency between predicted probabilities and empirical accuracy.
    \item \textbf{Agreement:} Cohen's Kappa coefficient ($\kappa$)~\cite{cohen1960kappa} to quantify the inter-rater reliability between the two judges.
\end{itemize}

\subsubsection{Hallucination Counts and Structural Integrity}
To provide a granular view of suppression beyond probability scores, we report the \textbf{Hallucination Sentence Count (HSC)}, calculated via strict keyword matching against our trigger dictionary (Table~\ref{tab:hsr_triggers}), to provide a granular count of removed hallucinations.

\subsubsection{Metric Decoupling and Structural Metrics}
To formally test our hypothesis that generic text-similarity metrics are decoupled from our task of hallucination suppression, we conduct a decoupling analysis.
For each generated report $y$, we compute its HSR and its similarity to the baseline $y^{(0)}$ using BERTScore~\cite{zhang2020bertscore}.
We perform a linear regression to predict a report's probability of being history-free, using both task-aligned and generic metrics as predictors while controlling for text length.
The results, presented in Table~\ref{tab:metric_decoupling}, confirm that both metric types are statistically significant, yet they capture different aspects of the text.

\begin{table}[h!]
    \centering
    \caption{Linear Regression validating the need for a decoupled evaluation.
    The dependent variable is the probability of a report being history-free.
    The statistical significance of both BERTScore and HSR confirms that while generic similarity is a factor, task-aligned metrics are indispensable for evaluating this specific failure mode.}
    \label{tab:metric_decoupling}
    \renewcommand{\arraystretch}{1.2}
    \begin{tabularx}{\columnwidth}{>{\raggedright\arraybackslash}X r r r r}
        \hline
        \hline
        \textbf{Predictor} & \textbf{Coefficient} & \textbf{Std. Error} & \textbf{t-statistic} & \textbf{$P>|t|$} \\
        \hline
        BERTScore           & -2.368 & 0.004 & -598.73 & $<$0.001 \\
        HSR (token-level)   & -1.379 & 0.004 & -334.36 & $<$0.001 \\
        Report Length       & -0.001 & 1.41e-06 & -825.70 & $<$0.001 \\
        Intercept           &  2.981 & 0.003 & 881.77 & $<$0.001 \\
        \hline
        \multicolumn{5}{l}{R-squared: 0.254, N = 2,792,822} \\
        \hline
        \hline
    \end{tabularx}
\end{table}

Motivated by this decoupling, we extend our evaluation beyond surface-level similarity. 
To assess whether the removal of historical content impacts the report's semantic structure, we introduce \textbf{RadGraph} (F1 on extracted clinical entities and relations)~\cite{jain2021radgraph} and \textbf{RadCliQ} (a composite metric of BLEU and CheXpert alignment)~\cite{yu2023radcliq}, alongside standard generation metrics such as BERTScore~\cite{zhang2019bertscore}. 
These metrics serve as auxiliary safety checks: stable scores indicate that the intervention preserves the relational structure of the clinical narrative despite the removal of comparison clauses.

\section{Results}
\label{sec:results}
Our experimental results are organized to first establish the efficacy of our method on a state-of-the-art model, utilizing a comprehensive benchmark across multiple injection strategies. We then analyze architectural dependencies and provide interpretability insights linking the ``positive-sum" performance to the promotion of visual grounding.


\subsection{Operating Point Selection}
We identify the optimal intervention configuration by applying the Dual-Objective Protocol (Sec.~\ref{ssec:exp_evaluation_metrics}), selecting the operating point that maximizes hallucination suppression subject to the clinical fidelity non-inferiority constraint.

\begin{table*}[t!]
\centering
\caption{\textbf{Comprehensive Performance Benchmark.} 
We report the optimal operating point for each strategy. 
``$\Delta$'' denotes the absolute change relative to the baseline.  \textcolor{blue}{\textbf{Blue}} highlights the primary reduction in hallucination probability for the optimal operating point. \textbf{Bold} values indicate the best performance per column.
Metrics include \textbf{HSR} (History-comparison Span Rate, measuring the proportion of text containing historical cues) and \textbf{FilBERT} (a detector for hallucinated prior comparisons, $\downarrow$).}
\label{tab:comprehensive_results}
\setlength{\tabcolsep}{3.5pt}
\renewcommand{\arraystretch}{1.1} 
\resizebox{\textwidth}{!}{%
\begin{tabular}{l l l c c c c c c c c c}
\toprule
\textbf{Model} & \textbf{Vector Source} & \textbf{Injection Strategy} & \textbf{$\lambda$} & \textbf{FilBERT} ($\downarrow$) & \textbf{$\Delta$} & \textbf{HSR} ($\downarrow$) & \textbf{$\Delta$} & \textbf{Macro-F1} ($\uparrow$) & \textbf{$\Delta$} & \textbf{RadGraph} & \textbf{RadCliQ} \\
\midrule
\multicolumn{12}{c}{\textbf{\textit{In-Domain Evaluation: MIMIC-CXR}}} \\
\midrule
\textbf{BiomedGPT} & -- & \textit{Baseline} & -- & 0.2373 & -- & 0.0081 & -- & 0.2242 & -- & \textbf{0.0035} & \textbf{0.0051} \\
 & SDIV (Ours) & GentleInject & -15.0 & 0.2350 & -0.0024 & 0.0087 & +0.0006 & 0.2913 & +0.0671 & -- & -- \\
 & SDIV (Ours) & ICV-Token & 0.0 & 0.2373 & +0.0000 & 0.0081 & +0.0000 & 0.2950 & +0.0707 & -- & -- \\
 & SDIV (Ours) & Global Injection (All) & -0.3 & 0.2311 & -0.0063 & \textbf{0.0046} & \textbf{-0.0035} & 0.3153 & +0.0911 & -- & -- \\
 & \textbf{SDIV (Ours)} & \textbf{SteerFair (Attn)} & \textbf{-0.3} & \textbf{0.1889} & \textcolor{blue}{\textbf{-0.0485}} & 0.0079 & -0.0002 & \textbf{0.3208} & \textbf{+0.0966} & 0.0026 & 0.0048 \\
 & Specific-50 & GentleInject & -10.0 & 0.2447 & +0.0073 & 0.0094 & +0.0013 & 0.1929 & -0.0313 & -- & -- \\
 & Specific-50 & SteerFair (Attn) & -0.2 & 0.2154 & -0.0219 & 0.0081 & -0.0000 & 0.2210 & -0.0032 & -- & -- \\
 & Global ICV & GentleInject & -20.0 & 0.2354 & -0.0019 & 0.0080 & -0.0002 & 0.1720 & -0.0522 & -- & -- \\
 & Global ICV & Global Injection (All) & -0.1 & 0.3173 & +0.0800 & 0.0092 & +0.0011 & 0.1907 & -0.0335 & -- & -- \\
\midrule
\textbf{VED Model} & -- & \textit{Baseline} & -- & 0.2703 & -- & 0.0111 & -- & 0.1769 & -- & 0.0049 & 0.0083 \\
 & SDIV (Ours) & GentleInject & -25.0 & 0.4600 & +0.1897 & \textbf{0.0000} & \textbf{-0.0111} & 0.1503 & -0.0266 & -- & -- \\
 & SDIV (Ours) & Global Injection (All) & -0.1 & 0.2887 & +0.0184 & 0.0112 & +0.0001 & 0.1706 & -0.0063 & -- & -- \\
 & SDIV (Ours) & SteerFair & -0.1 & 0.2903 & +0.0200 & 0.0113 & +0.0002 & 0.1730 & -0.0039 & \textbf{0.0485} & \textbf{0.0303} \\
 & \textbf{Specific-50} & \textbf{Global Injection (All)} & \textbf{-0.4} & \textbf{0.2220} & \textcolor{blue}{\textbf{-0.0483}} & 0.0170 & +0.0059 & \textbf{0.1858} & \textbf{+0.0089} & 0.0116 & 0.0118 \\
 & Global ICV & Global Injection (All) & -0.4 & 0.2335 & -0.0368 & 0.0256 & +0.0145 & 0.1757 & -0.0012 & 0.0117 & 0.0115 \\
\midrule
\textbf{LLaVA-Med} & -- & \textit{Baseline} & -- & 0.0505 & -- & 0.0000 & -- & \textbf{0.1837} & -- & 0.0000 & 0.0023 \\
 & SDIV (Ours) & GentleInject & -10.0 & \textbf{0.0505} & +0.0000 & \textbf{0.0000} & +0.0000 & \textbf{0.1837} & +0.0000 & -- & -- \\
 & SDIV (Ours) & Global Injection (All) & -0.2 & \textbf{0.0505} & -0.0000 & \textbf{0.0000} & +0.0000 & \textbf{0.1837} & +0.0000 & \textbf{0.0009} & \textbf{0.0028} \\
\midrule
\midrule
\multicolumn{12}{c}{\textbf{\textit{Zero-Shot Transfer: CheXpert Plus}}} \\
\midrule
\textbf{BiomedGPT} & -- & \textit{Baseline} & -- & 0.2230 & -- & 0.1576 & -- & 0.3030 & -- & \textbf{0.0011} & \textbf{0.0012} \\
 & SDIV (Ours) & GentleInject & -20.0 & 0.2957 & +0.0727 & \textbf{0.1259} & \textbf{-0.0317} & 0.2905 & -0.0125 & -- & -- \\
 & SDIV (Ours) & Global Injection (All) & -0.2 & 0.1629 & -0.0600 & 0.1376 & -0.0200 & 0.3076 & +0.0046 & -- & -- \\
 & \textbf{SDIV (Ours)} & \textbf{SteerFair (Attn)} & \textbf{-0.5} & \textbf{0.1532} & \textcolor{blue}{\textbf{-0.0697}} & 0.1720 & +0.0144 & \textbf{0.3234} & \textbf{+0.0204} & 0.0006 & \textbf{0.0012} \\
\midrule
\midrule
\multicolumn{12}{c}{\textbf{\textit{Zero-Shot Transfer: IU-Xray}}} \\
\midrule
\textbf{BiomedGPT} & -- & \textit{Baseline} & -- & 0.1980 & -- & 0.1663 & -- & 0.4115 & -- & 0.0062 & 0.0093 \\
 & SDIV (Ours) & GentleInject & -15.0 & 0.1747 & -0.0233 & 0.1225 & -0.0438 & 0.4000 & -0.0115 & -- & -- \\
 & \textbf{SDIV (Ours)} & \textbf{Global Injection (All)} & \textbf{-0.1} & \textbf{0.1241} & \textcolor{blue}{\textbf{-0.0739}} & \textbf{0.1195} & \textbf{-0.0468} & 0.4162 & +0.0047 & -- & -- \\
 & SDIV (Ours) & SteerFair (Attn) & -0.1 & 0.1881 & -0.0099 & 0.1443 & -0.0220 & \textbf{0.4335} & \textbf{+0.0220} & \textbf{0.0075} & \textbf{0.0122} \\
\bottomrule
\end{tabular}%
}
\end{table*}
\subsection{SDIV Delivers Clinically Aligned Suppression on BiomedGPT}
\label{ssec:results_biomedgpt}

Our primary finding is that on the advanced BiomedGPT model, SDIV effectively broke the ``suppression-fidelity'' trade-off.
A detailed quantitative comparison is provided in Table~\ref{tab:comprehensive_results}.

\paragraph{Failure of Naive Baselines}
The Global ICV baseline reveals the risks of entangled steering. As shown in the MIMIC-CXR section of Table~\ref{tab:comprehensive_results}, the \texttt{GentleInject} strategy with Global ICV ($\lambda=-20.0$) results in a negligible reduction in hallucinations (FilBERT 0.2354) but caused a catastrophic degradation in clinical fidelity, with Macro-F1 dropping from $0.2242$ to $0.1720$. This confirms that the global ``history" direction is heavily entangled with general clinical semantics; blindly suppressing it damages the model's core reasoning capabilities.

\paragraph{Clinical Safety}
It is crucial to correctly interpret the metric landscape. Traditional n-gram metrics or even structural graph metrics like RadGraph (0.0035 vs. 0.0026 in Table~\ref{tab:comprehensive_results}) are insufficient for detecting hallucinations, as historical comparisons often maintain valid grammatical and relational structure.
The decisive evidence for clinical safety is found in the CheXpert Macro-F1 scores.
As shown in Table~\ref{tab:comprehensive_results}, while naive baselines degraded F1 (e.g., Global ICV dropped to 0.1720), our SDIV intervention breaks this ceiling. The simultaneous improvement in FilBERT (0.2373 $\to$ 0.1889) and Macro-F1 (0.2242 $\to$ 0.3208) provided empirical proof that hallucination suppression does not have to come at the cost of diagnostic accuracy, validating the semantic precision of our orthogonal decomposition.
This divergence, characterized by a drop in FilBERT scores alongside an increase in F1, is the mathematical signature of surgical removal. It confirms that we are excising non-factual ``noise'' without damaging the ``signal'' of the patient's current pathology.

\paragraph{Success of SDIV via Targeted Attention Steering}
In contrast, our SDIV achieved a robust ``positive-sum" outcome. Crucially, our benchmark reveals that the injection strategy matters. While \texttt{Global Injection} (all-layer injection) improves F1, it provides limited suppression. The optimal strategy proved to be \textbf{\texttt{SteerFair (Attn)}}, which injects the vector specifically at the output of attention modules. At $\lambda=-0.3$, this configuration achieved the optimal trade-off, significantly reducing hallucinations while maximizing clinical fidelity.

This finding is mechanistically significant: it suggests that the ``hallucination signal" is most concentrated in the attention mechanism itself. By steering the attention outputs with our decoupled SDIV, we surgically suppressed the dominant 'historical comparison' prior. This disentanglement allows the model to surface latent clinical semantics that were previously overshadowed by statistical bias, without disrupting the subsequent feed-forward layers. 

\paragraph{Mechanistic Advantage of Orthogonalization}
The performance gap between Global ICV and SDIV provides empirical validation for our semantic orthogonality hypothesis.
As shown in Table~\ref{tab:comprehensive_results}, the Global ICV (derived via standard PCA) failed to distinguish between historical style and clinical content.
Specifically, under the \texttt{GentleInject} strategy, Global ICV caused a catastrophic degradation in clinical fidelity (Macro-F1 dropped by $0.0522$) while offering negligible suppression ($\Delta$FilBERT $\approx 0$).
This confirms that the principal component of the difference matrix $\mathbf{D}^{(m)}$ is highly correlated with essential clinical semantics; subtracting it aggressively removes valid medical findings along with the bias.

In contrast, our SDIV, constructed via $QR$-based semantic decomposition, successfully breaks this entanglement.
By projecting the historical direction onto a subspace orthogonal to the clinical findings, SDIV achieved a significant reduction in hallucinations ($\Delta$FilBERT $-0.0485$) while \emph{improving} clinical alignment (Macro-F1 increases by $+0.0966$).
This ``positive-sum'' outcome demonstrates that the $QR$ step functions as a safety filter, aiming to maximize the separation so that the intervention vector targets primarily the non-factual ``noise'' of historical comparison while preserving the ``signal'' of the patient's current condition.

\subsection{Cross-Dataset Generalization (Zero-Shot Transfer)}
\label{ssec:results_generalization}
A critical property of an interpretable latent steering vector is its universality. We derived the SDIV exclusively from the MIMIC-CXR training set. To test whether the ``historical comparison'' subspace is consistent across different hospital systems and reporting styles, we applied this MIMIC-derived vector directly to the CheXpert Plus~\cite{chambon2024chexpert} and IU-Xray~\cite{demner2016preparing} test sets without any fine-tuning (zero-shot transfer).

As detailed in the zero-shot transfer sections of Table~\ref{tab:comprehensive_results}, the MIMIC-derived SDIV transfers effectively without fine-tuning. 
On CheXpert Plus, the \texttt{SteerFair} strategy reduced the FilBERT score by 31.3\% ($\Delta = -0.0697$) while boosting Macro-F1 by 2.0\%. 
Similarly, on IU-Xray, the \texttt{Global Injection} strategy achieved a massive 37.3\% reduction in hallucination probability (FilBERT $0.1980 \to 0.1241$) while maintaining clinical fidelity.

These results suggest that the historical style bias exists as a distinct, transferable axis across different data distributions, and that our orthogonal decomposition captures this intrinsic feature rather than dataset-specific correlations.

\begin{table}[t]
\centering
\caption{\textbf{Hallucination Sentence Count (HSC).} Average hallucinated sentences per report across datasets, showing SDIV's reductions on foundation models.}
\label{tab:hallucination_removal_stats}
\resizebox{\columnwidth}{!}{%
\setlength{\tabcolsep}{3pt} 
\renewcommand{\arraystretch}{1.1}
\begin{tabular}{l l c c c c}
\toprule
\textbf{Dataset} & \textbf{Model} & \textbf{Base} & \textbf{SDIV} & \textbf{$\Delta$} & \textbf{Reduction (\%)} \\
\midrule
\textbf{MIMIC-CXR} (Source) & BiomedGPT & 0.22 & 0.16 & 0.06 & 26.2 \\
\midrule
\textbf{CheXpert Plus} (Zero-Shot) & BiomedGPT & 0.18 & 0.17 & 0.01 & 2.0 \\
 & VED & 1.80 & 1.27 & \textbf{0.53} & 40.9 \\
\midrule
\textbf{IU-Xray} (Zero-Shot) & BiomedGPT & 0.24 & 0.14 & 0.10 & 40.6 \\
\bottomrule
\end{tabular}%
}
\end{table}


\begin{table*}[t!]
\caption{\textbf{Qualitative Case Studies.} Comparison of generated reports for four representative test cases.
Metrics shown are \textbf{HSR} (History-comparison Span Rate, $\downarrow$), \textbf{FilBERT} (Hallucination Probability, $\downarrow$), and \textbf{F1} (CheXpert Macro-F1, $\uparrow$).
\halluc{Red} text denotes historical hallucinations; \fact{Green} text denotes valid clinical findings.
Note how SDIV surgically removes the red comparisons while maintaining or improving the green F1 scores.}
\label{tab:qualitative_analysis}
\centering
\small
\renewcommand{\arraystretch}{1.3}
\begin{tabularx}{\textwidth}{p{0.15\textwidth} X X X}
\toprule
\textbf{Metrics} & \textbf{Ground Truth} & \textbf{Baseline (No Intervention)} & \textbf{SDIV } \\
\midrule

\textbf{Case 1} \newline \textit{HSR}: 0.00 $\to$ 0.00 \newline \textit{FilBERT}: 2.00 $\to$ \textbf{0.00} \newline \textit{F1}: 0.86 $\to$ \textbf{0.95} & 
Pulmonary edema with bilateral pleural effusions, left greater than right. & 
ap chest \halluc{compared to [DATE]:} moderate - to - large left pleural effusion \halluc{has increased since [DATE]}. small right pleural effusion and mild - to - moderate pulmonary edema are \halluc{stable}. severe cardiomegaly is \halluc{chronic}. transvenous right atrial and right ventricular pacer leads in standard placements. no pneumothorax. & 
mild \fact{pulmonary edema} with \fact{bilateral pleural effusions}. \\
\midrule

\textbf{Case 2} \newline \textit{HSR}: 0.00 $\to$ 0.00 \newline \textit{FilBERT}: 2.00 $\to$ \textbf{0.00} \newline \textit{F1}: 0.84 $\to$ \textbf{0.87} & 
Diffuse bilateral opacities, most likely pulmonary edema, studies. However, superimposed infection cannot be excluded in the appropriate clinical setting. & 
\halluc{in comparison with the study of [DATE], there is little overall change.} again there are low lung volumes with enlargement of the cardiac silhouette and moderate pulmonary edema with bilateral pleural effusions and compressive basilar atelectasis. & 
\fact{mild pulmonary edema} with \fact{small bilateral pleural effusions}. \\
\midrule

\textbf{Case 3} \newline \textit{HSR}: 0.00 $\to$ 0.00 \newline \textit{FilBERT}: 2.00 $\to$ \textbf{0.00} \newline \textit{F1}: 0.84 $\to$ \textbf{0.88} & 
1. Right hilar mass, worrisome for neoplasm, warrants evaluation with a contrast enhanced chest ct. 2. Small bilateral pleural effusions. Notification : findings posted to the critical results dashboard on \_ \_ \_. & 
\halluc{in comparison with the study of [DATE], the patient has taken a better inspiration.} continued enlargement of the cardiac silhouette without definite vascular congestion or acute focal pneumonia. blunting of the costophrenic angles is \halluc{again seen}. & 
\fact{small bilateral pleural effusions}. \\
\midrule

\textbf{Case 4} \newline \textit{HSR}: 0.00 $\to$ 0.00 \newline \textit{FilBERT}: 3.00 $\to$ \textbf{0.00} \newline \textit{F1}: 0.84 $\to$ \textbf{0.85} & 
1. Free intraperitoneal air. \_ \_ \_ was paged to discuss this finding on \_ \_ \_ at 3 : 10 p. M. At the time of discovery. 2. Pulmonary edema. 3. Bibasilar atelectasis and small pleural effusions. & 
left picc line tip is at the level of mid svc. heart size and mediastinum are \halluc{stable}. bilateral pleural effusions are large, \halluc{unchanged}. there is interval progression of pulmonary edema. & 
1. left - sided picc line terminates in the left brachiocephalic vein. 2. \fact{small bilateral pleural effusions}. \\

\bottomrule
\end{tabularx}
\end{table*}

This surgical precision is confirmed in Table~\ref{tab:qualitative_analysis}.
In Case 1, the Baseline model hallucinates an explicit comparison (``has increased since [DATE]''). Our SDIV intervention surgically removes this temporal reference while preserving the core pathology (``pulmonary edema''). 
As shown in the table, the Hallucination Span Rate (HSR) drops to \textbf{0.00} while the CheXpert Macro-F1 score improves, quantitatively verifying that the excised information was indeed noise rather than clinical signal. 
Similarly, in Case 2, the Baseline's preamble is removed without degrading clinical fidelity metrics, confirming the intervention's specificity to the non-factual ``noise'' of history.

\subsection{Dependencies: Architecture and Vector Specificity}

The VED and LLaVA-Med sections of Table~\ref{tab:comprehensive_results} reveal critical boundaries regarding model architecture and vector specificity.

\paragraph{Data Specificity Sensitivity (VED Model)}
On the VED model, results highlight the primacy of data specificity.
The automated SDIV ($\lambda=-0.1$) struggled to suppress hallucinations on this architecture.
However, the \textbf{Curated-Specific-50} vector (derived from manually curated, minimal-edit pairs) achieved the best suppression in the group (FilBERT 0.2220), acting as an empirical ``Oracle" or upper bound for specificity.
Interestingly, the Global ICV performs better than SDIV here. This suggests that smaller, less robust models like VED may struggle with the nuanced semantic directions extracted by SDIV, responding better to the ``brute force" signal of Global ICV or the perfect specificity of Specific-50.

\paragraph{Architectural Boundary: The Scope of Linear Steering}
The distinct behavior observed on LLaVA-Med ($\Delta \approx 0$ across all metrics) illuminates a critical scope condition for our framework. Unlike BiomedGPT, which integrates vision via deep, layer-wise cross-attention, LLaVA-Med adopts a visual-prefix architecture where image features act as initial prompts. This structural contrast suggests that effective linear steering of cross-modal hallucinations benefits significantly from a persistent, explicit interface between modalities. In prefix-based architectures, the ``comparison'' bias appears to become diffused throughout the global self-attention stream. This results in a more complex, non-linear entanglement that may require interventions beyond the linear subspace decomposition targeted by SDLS, marking an important avenue for future generalized steering research.

\begin{figure*}[t!]
    \centering
    \subfloat[``...standard positioning of \fact{endotracheal} and enteric tubes...'']{
        \includegraphics[height=3.5cm, keepaspectratio]{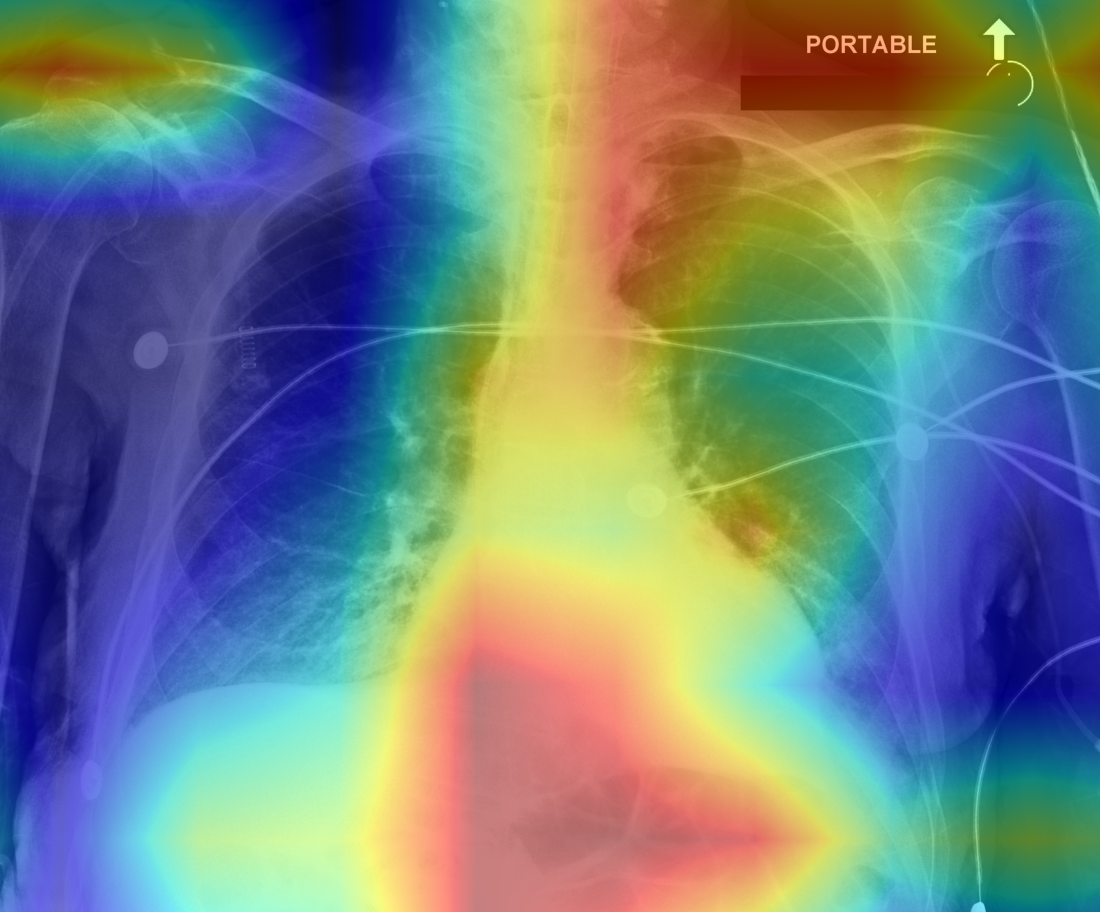} 
        \label{fig:attn_endotracheal}
    }
    \hfill
    \subfloat[``...no acute \fact{cardiopulmonary} abnormality seen...'']{
        \includegraphics[height=3.5cm, keepaspectratio]{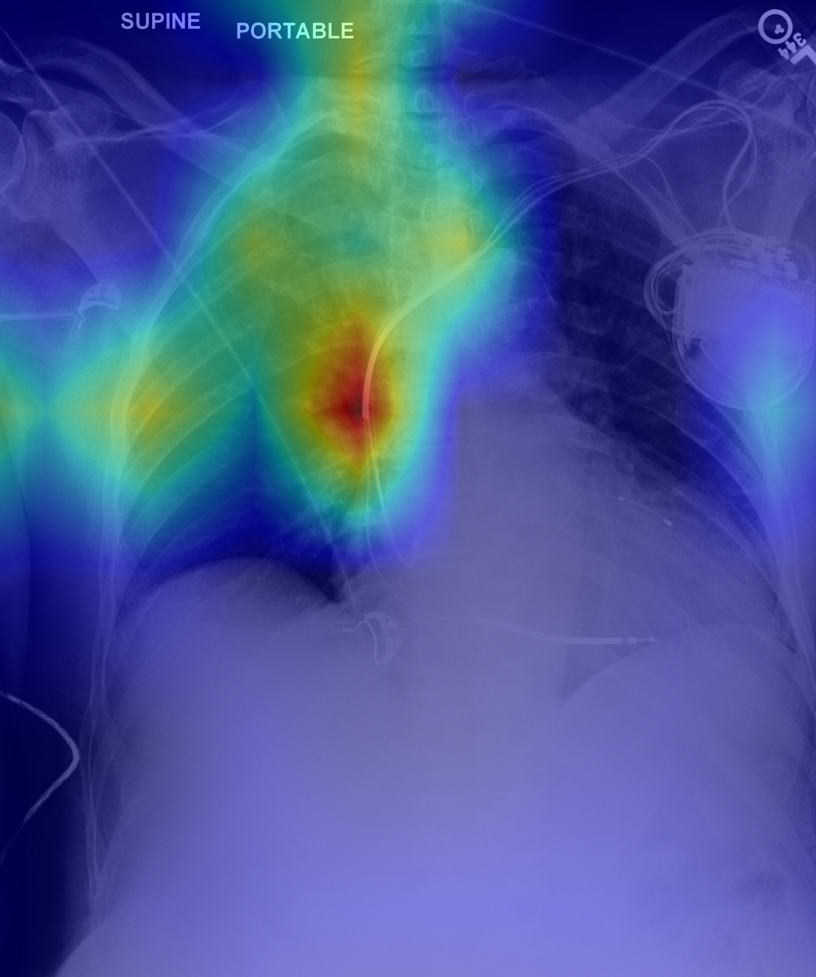}
        \label{fig:attn_cardiopulmonary}
    }
    \hfill
    \subfloat[``...support devices are \halluc{unchanged} from prior...'']{
        \includegraphics[height=3.5cm, keepaspectratio]{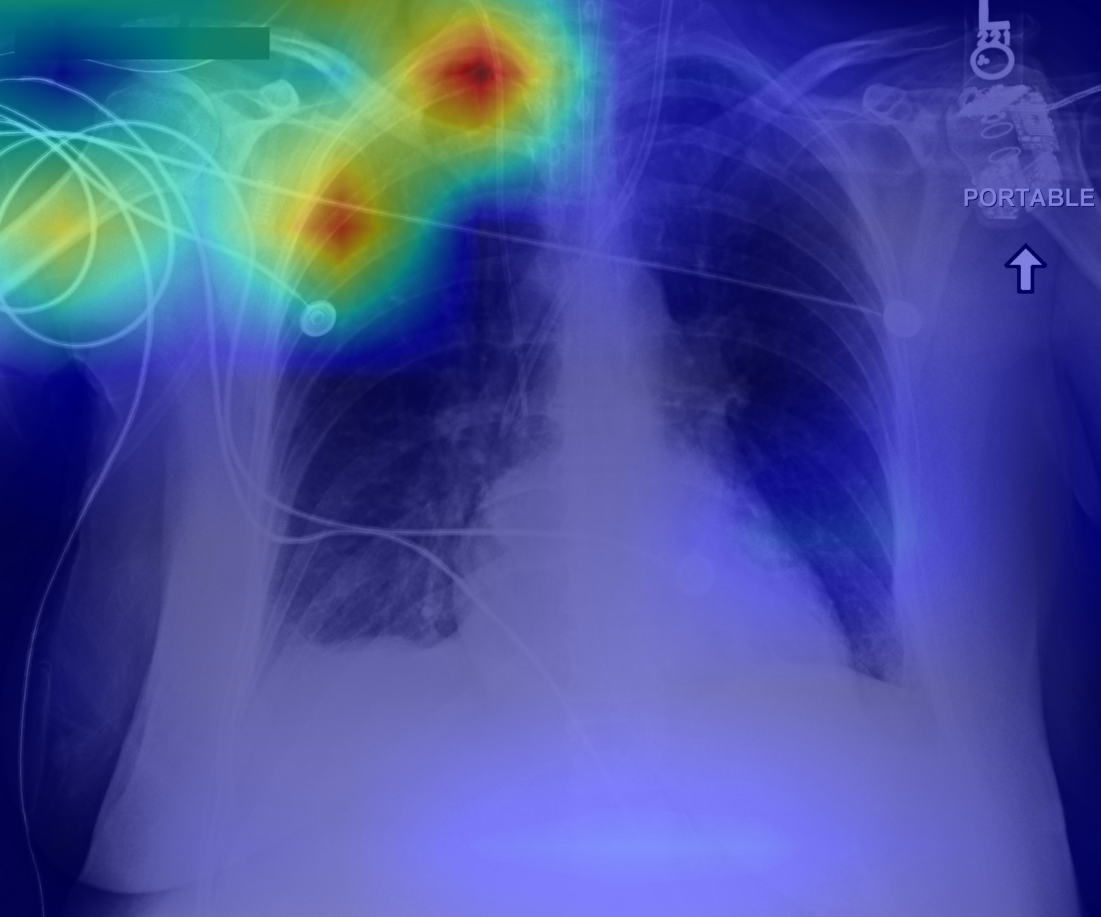}
        \label{fig:attn_unchanged}
    }
    \hfill
    \subfloat[``...cardiomediastinal silhouette is \halluc{stable} from...'']{
        \includegraphics[height=3.5cm, keepaspectratio]{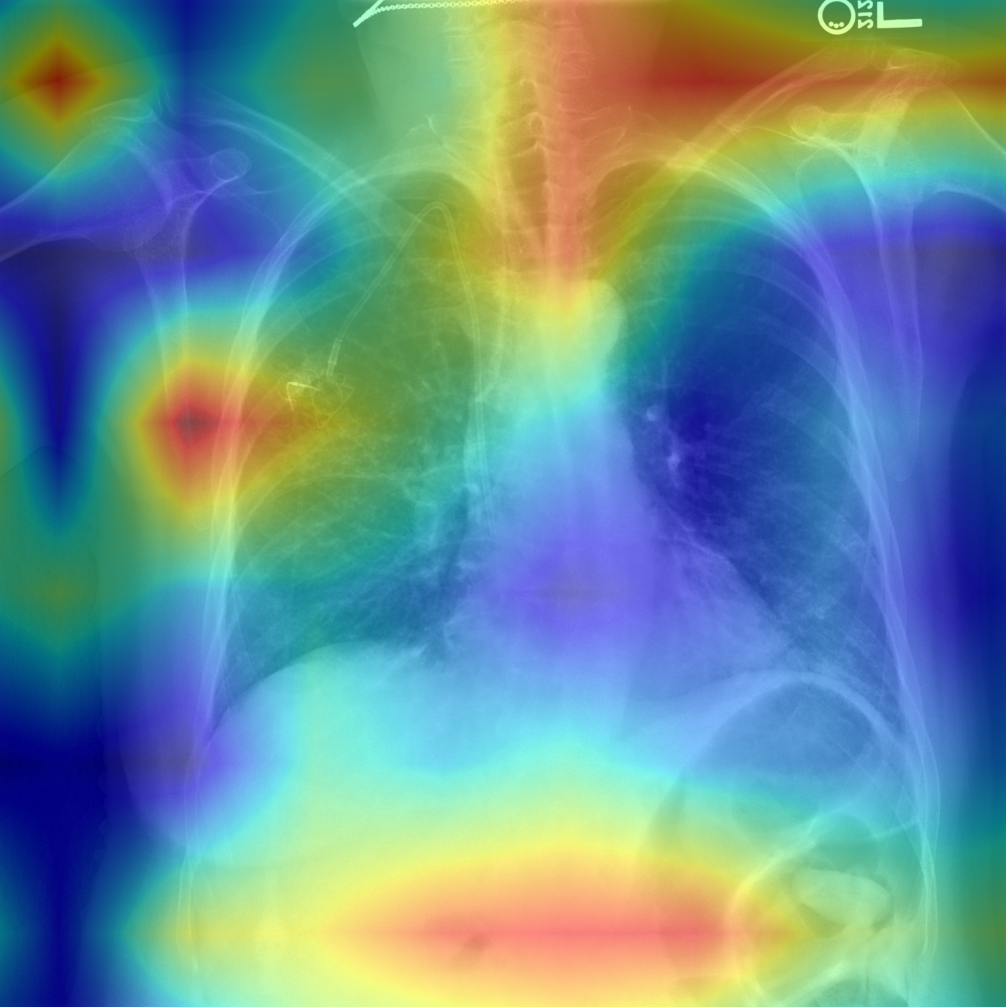}
        \label{fig:attn_stable}
    }

    \caption{\textbf{Qualitative Visualization of Visual Grounding.} 
    We visualize the cross-attention maps during the generation of specific target tokens.
    \textbf{(a, b) Factual terms:} For grounded terms like \fact{``endotracheal''} or \fact{``cardiopulmonary''}, the model's attention is strongly focused on the relevant anatomical regions (trachea, heart), indicating robust visual grounding.
    \textbf{(c, d) Hallucinations:} For historical comparison terms like \halluc{``unchanged''} or \halluc{``stable''}, the attention map is diffuse and failed to focus on meaningful visual features, confirming that these words are driven by language priors rather than visual evidence.}
    \label{fig:attention_heatmaps}
\end{figure*}
\subsection{Qualitative Analysis: Visual Grounding vs. Language Priors}
\label{ssec:results_mechanism}
To understand the underlying cause of prior-comparison hallucinations, we conduct a direct analysis of the decoder's cross-attention mechanism.
As shown in Fig.~\ref{fig:attention_heatmaps}, a clear dichotomy emerges.
For factual terms (Fig.~\ref{fig:attention_heatmaps}a-b), such as ``endotracheal'' and ``cardiopulmonary'', the model's attention is tightly localized on the corresponding physical structures (the trachea and cardiac silhouette, respectively).
Conversely, for historical terms (Fig.~\ref{fig:attention_heatmaps}c-d) like ``unchanged'' and ``stable'', the attention maps are diffuse, focusing on background noise or support devices rather than relevant anatomy.
This visual decoupling suggests that prior-comparison hallucinations are driven by internal language priors that override visual evidence.
\begin{figure}[h!]
    \centering
    \includegraphics[width=0.9\columnwidth]{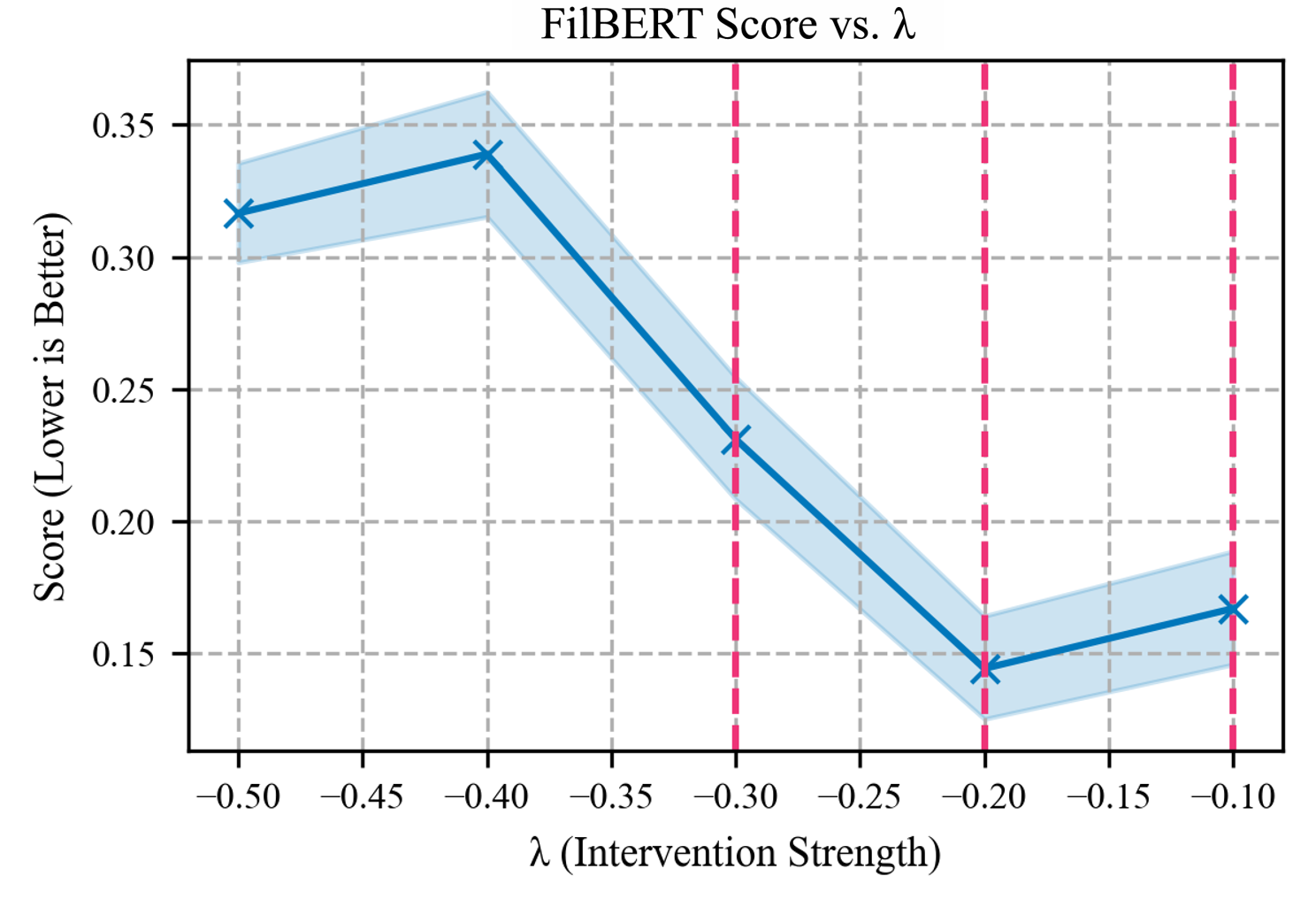}
    \caption{Dose-response curve showing FilBERT Score as a function of intervention strength ($\lambda$) for SDIV on BiomedGPT. The curve demonstrates a clear optimal intervention zone, highlighting the controllability of the method.}
    \label{fig:dose_response_curve}
\end{figure}
Additionally, Fig.~\ref{fig:dose_response_curve} illustrates the ``dose-response'' relationship for SDIV. The emergence of a clear optimal zone (e.g., $\lambda \approx -0.2$) confirms that the intervention provides a predictable, controllable steering mechanism rather than random perturbation.

\subsection{Robustness and Specificity}
\begin{figure}[h!]
    \centering
    \includegraphics[width=0.9\columnwidth]{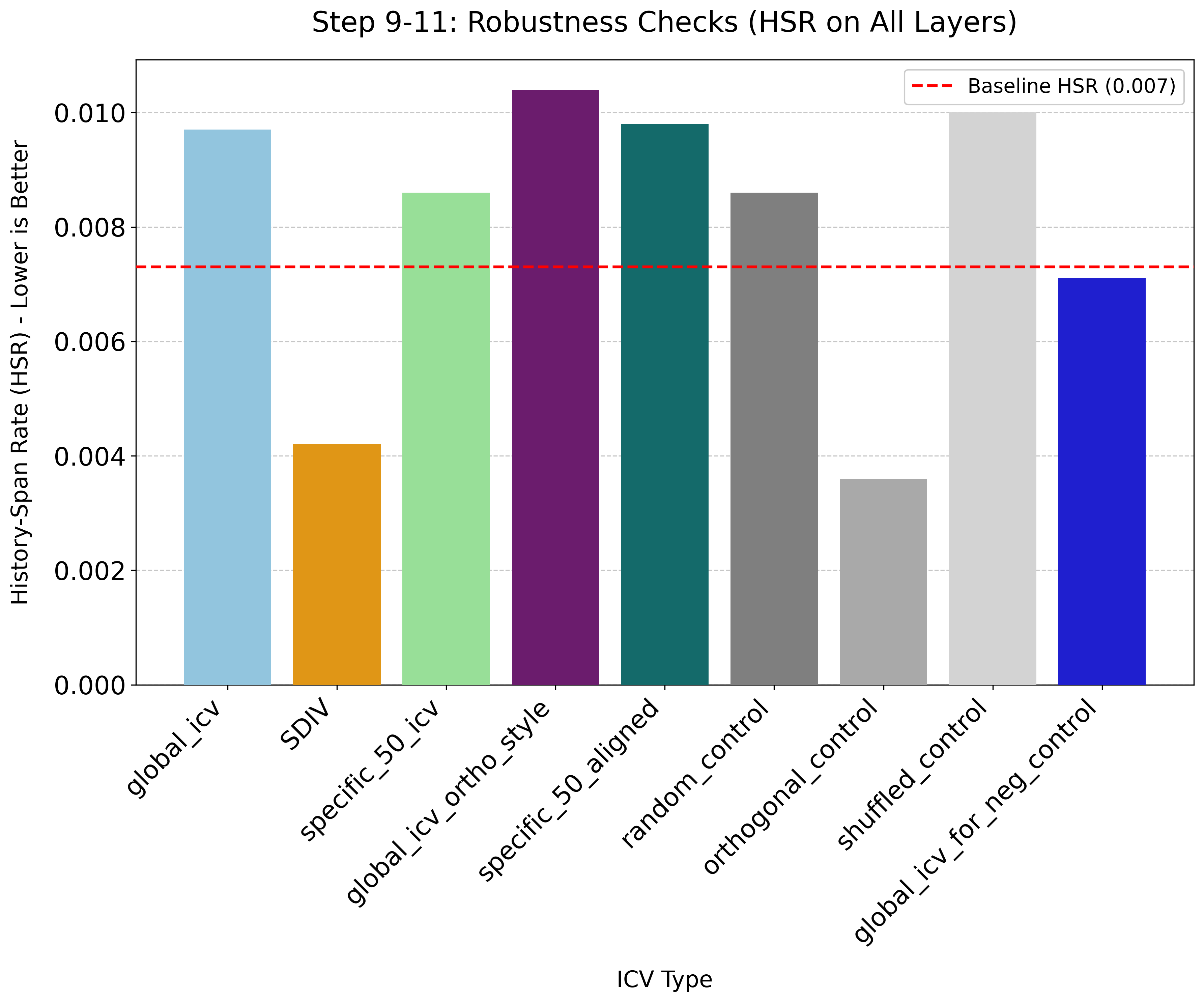}
    \caption{\textbf{Robustness and specificity checks.} Standard negative controls (random, shuffled) show no suppression effect. Note that the style-orthogonalized ICV (orange dashed) underperforms the original baseline.}
    \label{fig:robustness_checks}
\end{figure}

Finally, we leverage negative controls to further justify the necessity of our SDLS framework. As shown in Fig.~\ref{fig:robustness_checks}, standard negative controls (random, shuffled, orthogonal vectors) have no suppression effect, confirming specificity.
Most importantly, the \texttt{global\_icv\_ortho\_style} vector—an attempt to remove writing style via simple linear projection—performs \emph{worse} than the original Global ICV.
This negative result is a key finding: it proves that historical semantics and writing style are statistically entangled in the model's latent space. A simple subspace subtraction cannot separate them without damaging the semantic signal. This empirically justifies our use of LLM-based decomposition, which provides cleaner semantic supervision than raw differences and $QR$ orthogonalization, which together succeed in disentangling these factors where linear projection failed.

\section{Conclusion and Discussion}
\label{sec:conclusion}
This work introduced SDLS, a training-free, inference-time framework for suppressing prior-comparison hallucinations in radiology report generation. Our systematic investigation, centered on the construction of the SDIV, yielded three principal contributions. First, we demonstrated that on an advanced cross-attention model (BiomedGPT), SDIV can achieve a ``positive-sum" outcome, effectively reducing historical references while simultaneously improving the clinical fidelity of the generated reports. This success was maximized not by global injection, but by leveraging the cross-attention structure, verifying that the hallucination signal is concentrated in the cross-modal interface. Second, through comparative analysis, we identified critical dependencies: a persistent cross-attention mechanism is a key architectural prerequisite, and the semantic decoupling of the contrastive data is paramount. Third, we provided a direct mechanistic explanation, showing via attention analysis that such hallucinations are correlated with a failure of visual grounding. SDIV addresses this by neutralizing the statistical language priors that exacerbate this drift. This intervention creates a latent environment where factual, image-grounded descriptions prevail.

Our findings offer a viable framework for deploying representation steering techniques in a clinical context. Rather than seeking a single ``best'' method, we advocate for a paradigm of operating point selection. Here, the intervention strength is calibrated to maximize hallucination suppression, subject to a strict non-inferiority constraint on clinical fidelity.
Furthermore, the failure of the PCA-based Global ICV to separate historical semantics from clinical content empirically validates our core hypothesis: these factors are entangled in the raw latent space, necessitating the usage of our LLM-based decomposition and $QR$ orthogonalization to achieve precise, safe control.

This study has limitations that highlight important avenues for future research. First, our findings highlight foundational dependencies on data and model capacity. While SDIV excels on the robust BiomedGPT, the simpler VED model showed a preference for the ``brute force" signal of Global ICV or the ``oracle" specificity of \texttt{Curated-Specific-50}. This suggests that automated orthogonalization requires a model latent space sufficiently rich to represent the nuanced directions SDIV extracts. Second, the intervention exhibits hyperparameter sensitivity. Success relies on careful tuning of the intervention strength ($\lambda$) and the injection locus; as our results showed, targeting specific attention layers (\texttt{SteerFair}) yields significantly better trade-offs than indiscriminate all-layer injection (\texttt{Global Injection}). Third, from a mechanistic standpoint, while our analysis establishes a strong correlation between attention patterns and hallucinations, future work requires causal mediation analysis to fully isolate the specific attention heads responsible.

Furthermore, the successful application of our MIMIC-derived vectors to unseen datasets (CheXpert Plus, IU-Xray) demonstrates the broad transferability of the identified historical style subspace across diverse clinical settings\cite{xiao2024visual}. This capability aligns with the broader goals of generalized zero-shot learning~\cite{xian2019zero}, suggesting that SDLS can scale to diverse clinical environments without local retraining.

Looking forward, future work should address the dependency on text pairs by developing visually-grounded intervention vectors learned directly from image features. To improve usability, research could focus on automated methods for identifying the optimal operating point, mitigating the need for grid searches. Most importantly, future studies must move towards human-in-the-loop evaluation, assessing how SDIV-corrected reports impact radiologists' workflow and diagnostic confidence. Ultimately, SDLS reframes hallucination control from a black-box suppression task to a geometric disentanglement problem. By mathematically isolating the residual artifacts of historical priors from the visual manifold, we enable models to speak clearly about the present without being haunted by the statistics of the past. This work paves the way for a new class of transparent, inference-time steering mechanisms essential for trustworthy clinical AI.


\bibliographystyle{IEEEtran}
\bibliography{references}

\end{document}